\documentclass[10pt,journal,compsoc]{IEEEtran}
\ifCLASSOPTIONcompsoc
  \usepackage[nocompress]{cite}
\else
  \usepackage{cite}
\fi
\ifCLASSINFOpdf
\else
\fi
\usepackage{microtype}
\usepackage{graphicx}
\usepackage{subfigure}
\usepackage{booktabs} 
\usepackage{amsmath}
\usepackage{multirow}
\usepackage{pdfpages}
\usepackage{comment}
\usepackage[numbers]{natbib}
\usepackage{amsfonts}
\usepackage{hyperref}
\usepackage{times}

\begin{document}
%
\title{Convolution-enhanced Evolving Attention Networks}
%
%
%
%

\author{Yujing Wang, Yaming Yang, Zhuo Li, Jiangang Bai, Mingliang Zhang, \\Xiangtai Li, Jing Yu, Ce Zhang, Gao Huang, Yunhai Tong
\IEEEcompsocitemizethanks{\IEEEcompsocthanksitem Yujing Wang, Yaming Yang, Jiangang Bai, Mingliang Zhang, Xiangtai Li, Yunhai Tong are with Key Laboratory of Machine Perception, MOE, School of Intelligence Science and Technology, Peking University. E-mail: \{yujwang, pku\_bjg, zml24, lxtpku, yhtong\}@pku.edu.cn; yamingyang@stu.pku.edu.cn.\protect
\IEEEcompsocthanksitem Zhuo Li is with Department of Computer Science $\&$ Technology, Engineering Research Center of Microprocessor $\&$ System, Peking University. Email: lizhuo@stu.pku.edu.cn.\protect
\IEEEcompsocthanksitem Jing Yu is with Institute of Information Engineering, Chinese Academy of Sciences. E-mail: yujing02@iie.ac.cn.\protect
\IEEEcompsocthanksitem Ce Zhang is with ETH Z\"{u}rich. E-mail: ce.zhang@inf.ethz.ch.\protect
\IEEEcompsocthanksitem Gao Huang is with the Department of Automation, Tsinghua University. E-mail: gaohuang@tsinghua.edu.cn.\protect
\IEEEcompsocthanksitem
Yujing Wang and Yaming Yang contribute equally to this work. Yujing Wang is the corresponding author. 
}
\thanks{Manuscript received March 29, 2022; revised October 8, 2022.}}

%
%

\markboth{Journal of \LaTeX\ Class Files,~Vol.~14, No.~8, August~2015}%
{Shell \MakeLowercase{\textit{et al.}}: Bare Demo of IEEEtran.cls for Computer Society Journals}
%



\IEEEtitleabstractindextext{%
\begin{abstract}
Attention-based neural networks, such as Transformers, have become ubiquitous in numerous applications, including computer vision, natural language processing, and time-series analysis. In all kinds of attention networks, the attention maps are crucial as they encode semantic dependencies between input tokens. However, most existing attention networks perform modeling or reasoning based on \textit{representations}, wherein the \textit{attention maps} of different layers are learned separately \textit{without} explicit interactions. 
In this paper, we propose a \textit{novel} and \textit{generic} evolving attention mechanism, which directly models the \textit{evolution} of \textit{inter-token relationships} through a chain of residual convolutional modules.
The major motivations are twofold. On the one hand, the attention maps in different layers share transferable knowledge, thus adding a residual connection can facilitate the information flow of inter-token relationships across layers. On the other hand, there is naturally an evolutionary trend among attention maps at different abstraction levels, so it is beneficial to exploit a dedicated convolution-based module to capture this process.
Equipped with the proposed mechanism, the convolution-enhanced evolving attention networks achieve superior performance in various applications, including time-series representation, natural language understanding, machine translation, and image classification. Especially on time-series representation tasks, Evolving Attention-enhanced Dilated Convolutional (EA-DC-) Transformer outperforms state-of-the-art models significantly, achieving an average of 17\% improvement compared to the best SOTA. To the best of our knowledge, this is the first work that explicitly models the layer-wise evolution of attention maps. Our implementation is available at \textit{\url{https://github.com/pkuyym/EvolvingAttention}}.
\end{abstract}

}

\maketitle

\IEEEdisplaynontitleabstractindextext

%
\IEEEpeerreviewmaketitle

\IEEEraisesectionheading{\section{Introduction}\label{sec:introduction}}

\IEEEPARstart{A}{ttention}-based neural networks have achieved plentiful successes in various applications, such as natural language understanding~\cite{devlin2018bert}, image generation~\cite{parmar2018image} and time-series forecasting~\cite{li2019enhancing}. Most notably, the attention-based architecture, Transformer~\cite{vaswani2017attention}, has become ubiquitous in sequential modeling and established state-of-the-art performances on numerous tasks in computer vision, natural language, and time-series domains~\cite{dosovitskiy2020image,liu2019roberta,zerveas2021transformer}. The performance of an attention-based neural network mainly depends on its capability of inducing reasonable attention patterns among input tokens. However, as illustrated by some previous works~\cite{tang2018self,jain2019attention}, the attention maps captured by vanilla attention layers are not always effective and explainable. To cope with this problem, recent efforts concatenated self-attentions with convolutional layers to obtain better image or text representations~\cite{bello2019attention,wu2020lite}. These works mostly focus on improving representations, yet the attention map itself has not been ameliorated.
Here we raise a fundamental question, {\em can we improve the modeling of inter-token dependencies through dedicated architecture designs?} In this paper, we will address it by proposing a novel and generic \textit{evolving attention} mechanism that improves the quality and explainability of attention maps.

\begin{figure*}[t]
	\centering
	\begin{minipage}[t]{0.6\linewidth} 
        \includegraphics[width=1.0\linewidth]{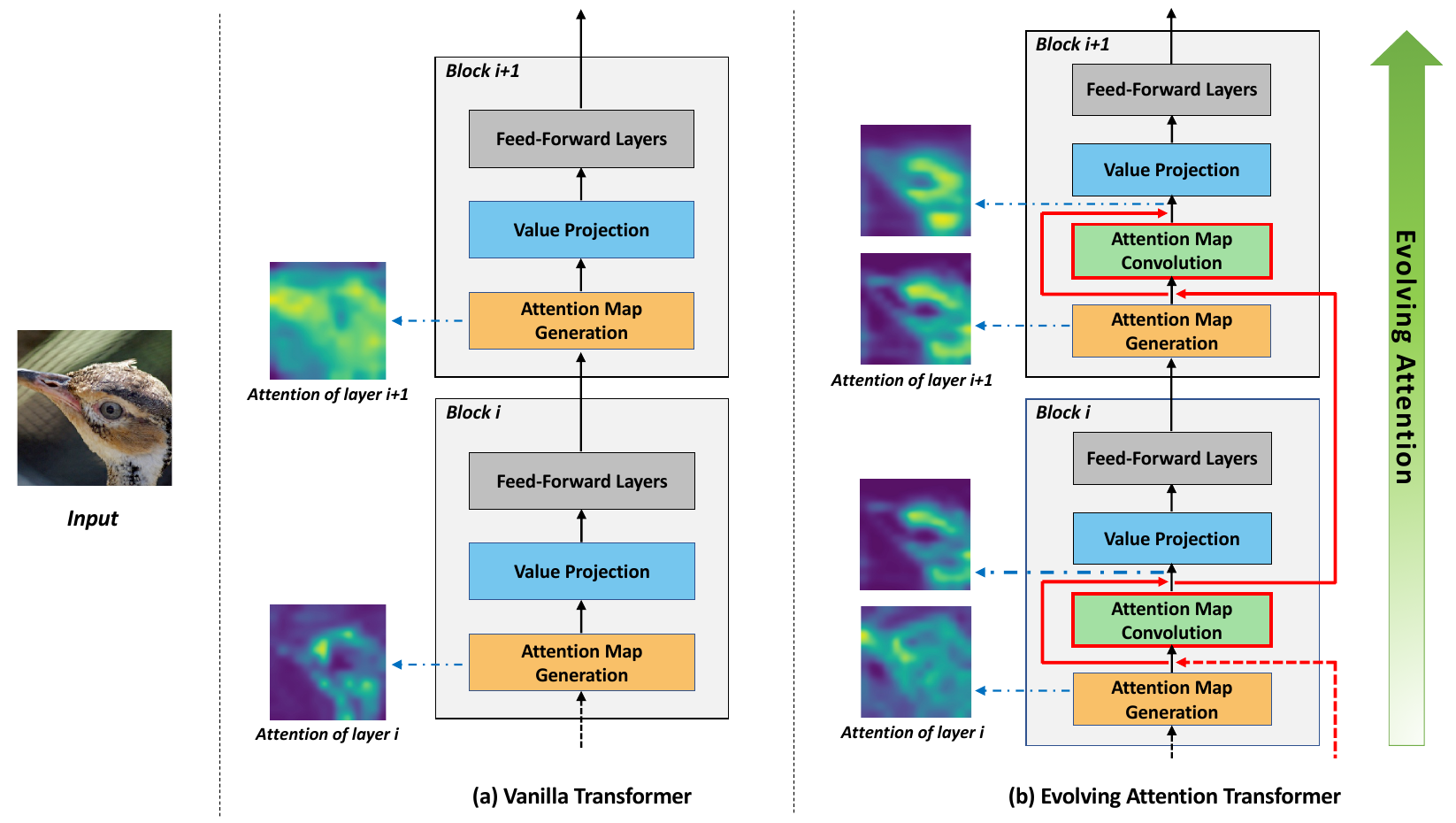}
    \end{minipage}
    \begin{minipage}[t]{0.36\linewidth} 
    \includegraphics[width=0.85\linewidth]{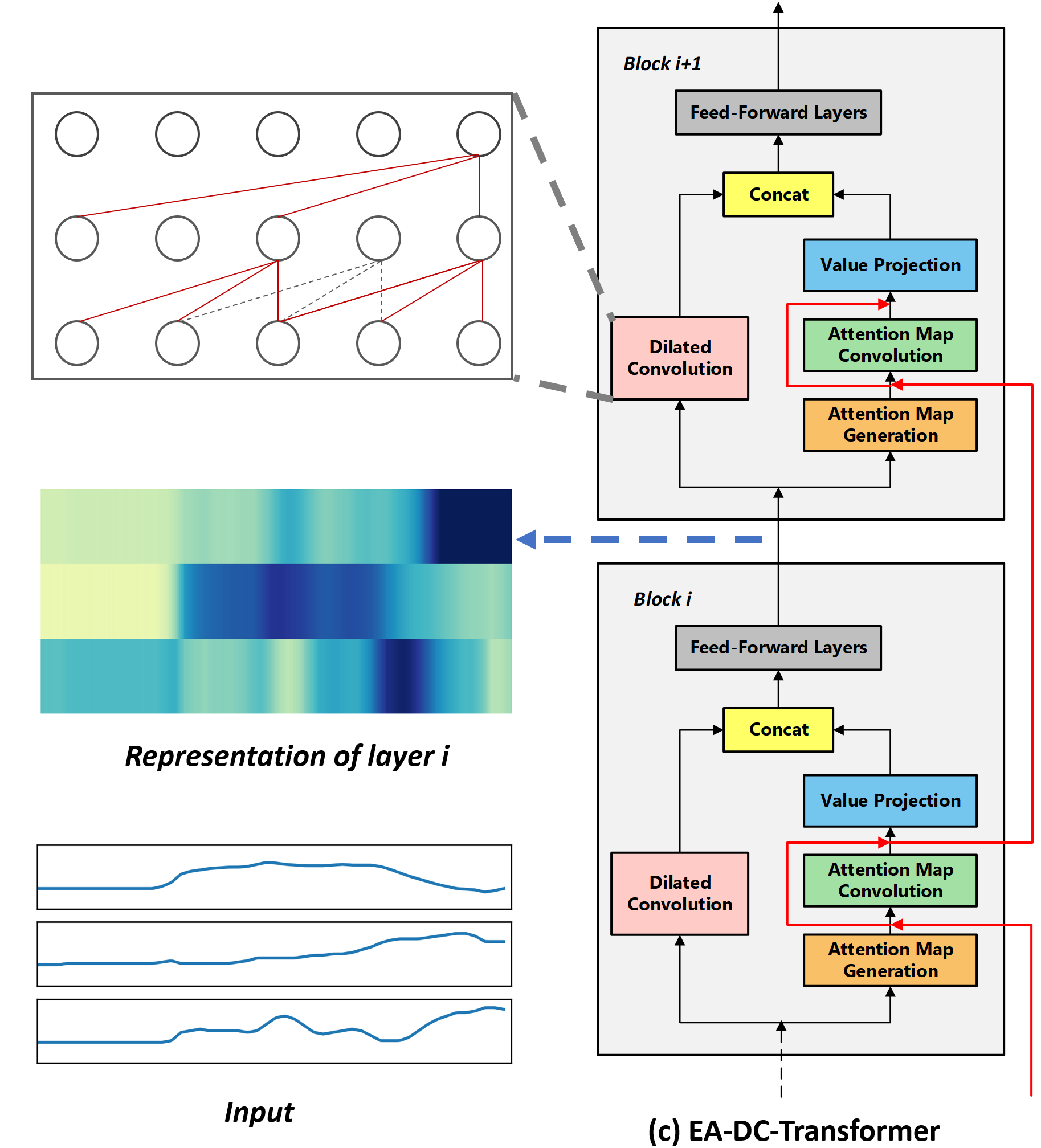}
    \end{minipage}
    \caption{Overview. Figure 1(a) shows a standard Transformer network, where we omit layer norms and skip connections in the figure for brevity; Figure 1(b) shows an Evolving Attention-enhanced Transformer, where the red lines denote residual connections, with exemplar attention maps from the 17th and 18th blocks presented. As shown by the case, vanilla transformer obtains very vague attention maps at the 18th block. Instead, evolving attention-enhanced transformer generates reasonable attention maps, and there is a clear evolutionary trend between adjacent layers. Figure 1(c) illustrates the network architecture of Evolving Attention-enhanced Dilated Convolutional (EA-DC-) Transformer for time-series representation.}
    \label{fig:overview}
\end{figure*}

Most existing attention neural networks conduct modeling based on \textit{input representations}, while the attention maps in each layer are learned \textit{without} explicit interactions, resulting in a bad generalization ability. As shown by a preliminary study on ImageNet classification, the attention maps learned by vanilla attention networks are sometimes ineffective and unexplainable. Figure \ref{fig:overview} visualizes two attention maps captured in the 17-th and 18-th layers, respectively, for the ImageNet classification task, wherein the corresponding attention map for the 18-th layer is rather vague. 
Our motivation is to design a dedicated module to directly capture the \textit{evolution of inter-token dependencies} based on attention maps. Thus, we directly bridge the attention maps from adjacent layers to facilitate information flow. Furthermore, we adopt additional convolutional layers to learn the evolution trend of attention patterns. This inductive bias will emphasize local details and induce better inter-token relationships by reasoning explicitly on previous attention maps.

To this end, we propose Convolution-enhanced Evolving Attention Networks\footnote{A preliminary version of this paper is published at ICML 2021~\cite{wang2021evolving}} (CEANs), which stand for a group of models that guides the learning of inter-token relationships via a chain of residual convolution layers over attention maps. In each layer, the CEAN model takes all attention maps generated by the previous layer as a multi-channel image. Then, with 2D convolutions over that multi-channel image, the attention maps for the current layer can be efficiently evolved from the ones of previous layers. As such, the generic knowledge of inter-token dependencies is shared across all blocks, while for each layer, the attention maps can be adapted to the appropriate abstraction level. As shown by the exemplar case in Figure \ref{fig:overview}(b), the attention maps learned by convolution-enhanced evolving attention networks have correctly highlighted the structure of bird with the assistance of this mechanism.
Instead, vanilla attention networks without evolutionary modules sometimes induce very vague attention patterns (see Figure \ref{fig:overview}(a)). Besides, the proposed mechanism is general and not restricted to the image-related applications. The inter-token and inter-timestamp relationships are also crucial for natural language and time series, and the proposed evolving attention mechanism can be widely applied in these two domains. Also, it has potential impacts in other scenarios, such as graph attention networks~\cite{velivckovic2018graph}.


\begin{figure}[t]
	\centering
        \includegraphics[width=0.9\linewidth]{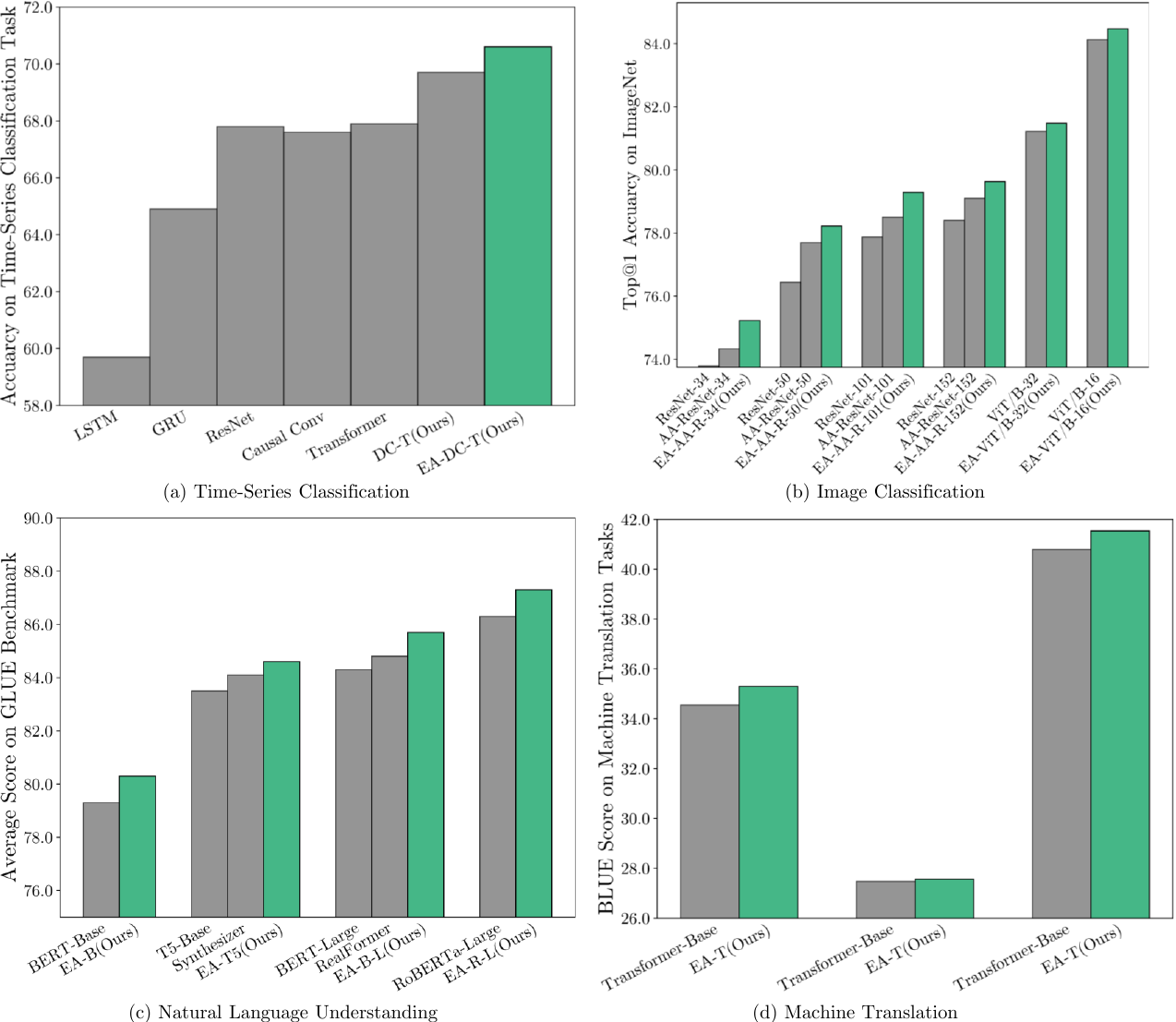}
         \caption{Empirical results on various tasks of different modalities: (a) Time-Series Classification, (b) Image Classification, (c) Natural Language Understanding, and (d) Machine Translation. In the figure, gray bars indicate baseline models, and the highlighted green bars indicate models equipped with convolution-enhanced evolving attention mechanism.}
        \label{fig:imagenet}
\end{figure}

There are multiple instantiations of convolution-enhanced evolving attention networks based on different kinds of backbone models. First, we propose a new backbone model for time-series representation, named Dilated Convolutional (DC-) Transformer. It integrates dilated convolution into the transformer network, which introduces effective inductive bias to assist the model to cope with time series of varying and potentially high lengths. Furthermore, we apply the evolving attention mechanism to enhance the proposed backbone, resulting in the EA-DC-Transformer architecture. As illustrated in Figure \ref{fig:imagenet}(a), DC-Transformer itself has strong performance, ranking in the first place among existing SOTAs and outperforming the best one by an average of 13\% on all time-series representation datasets.
Moreover, the integration of convolution-enhanced evolving attention mechanism further improves the performance of DC-Transformer by a large margin (3.5\% lifts in average).

In other domains, we adopt the commonly used state-of-the-art backbone networks, and add a prefix ``EA" to denote the corresponding network architecture augmented by the evolving attention mechanism. Specifically, we utilize transformer for machine translation, BERT~\cite{devlin2018bert}, RoBERTa~\cite{liu2019roberta} and T5~\cite{raffel2019exploring} for natural language understanding and two attention neural networks for image classification, including Attention Augmented (AA-) ResNet~\cite{bello2019attention} and Vision transformer (ViT)~\cite{dosovitskiy2020image}. We leverage proper convolution masks in the encoder-decoder and decoder networks so that the evolving attention mechanism is also applicable to sequence-to-sequence tasks like machine translation. 
On image classification (Figure \ref{fig:imagenet}(b)), EA-enhanced models consistently improve the accuracy of both AA-ResNet and ViT for different model capacities.
We also prove the effectiveness of evolving attention for various pre-trained language models (Figure \ref{fig:imagenet}(c)). In addition, the evaluation on neural machine translation tasks (Figure \ref{fig:imagenet}(d)) indicates that the proposed evolving attention mechanism is not only feasible for encoders, but also helpful for both encoder-decoder and decoder networks.

The contributions of this paper are highlighted as follows.
\begin{itemize}
    \item 
    We propose novel convolution-enhanced evolving attention networks augmented by a chain of residual convolutional modules. To the best of our knowledge, this is the first work that considers attention maps as multi-channel images and leverages convolution-based models for pattern extraction and layer-wise evolution. The proposed convolution-based evolving attention mechanism is \textit{generic} and can be applied to most kinds of -- if not all -- attention networks, which sheds new light on the traditional attention mechanism.
    \item
    We propose a novel backbone model, namely Dilated Convolutional (DC-) Transformer for time-series representation. It combines the benefits of dilated convolution and transformer networks, introducing plausible inductive bias for time series with varying lengths and achieving significant performance improvement over state-of-the-art time-series representation models. 
    \item 
    Extensive experiments demonstrate the effectiveness of evolving attention in various tasks in time series, natural language, and computer vision domains. As illustrated by our extensive analyses from the qualitative and quantitative perspectives, both residual connections and convolutional inductive bias are beneficial to generate more effective and explainable attention maps. 
\end{itemize}
\section{Related Work}
The idea of using attention in neural networks is first introduced by \cite{bahdanau2014neural}, which utilizes a context vector to align the source and target inputs in neural machine translation. The context vector preserves all representations of input tokens, while the current output token will attend to a certain part of the input sequence during the decoding process. The attention mechanism has achieved great success in computer vision~\cite{parmar2018image, parmar2019stand}, natural language~\cite{vaswani2017attention, devlin2018bert}, and time-series~\cite{zerveas2021transformer} domains. 

Luong \textit{et al.}~\cite{luong2015effective}
summarize various types of attentions, including dot-product, multiplicative, additive, and concatenative ones. Based on the dot-product attention mechanism, Vaswani \textit{et al.}~\cite{vaswani2017attention} propose the Transformer architecture for neural machine translation, which is then widely applied to numerous tasks in multiple domains~\cite{han2020survey}. In the following subsections, we will first present existing studies and works on attention maps. Then, we introduce the applications of attention neural networks in time-series representation, computer vision, and natural language processing, respectively.

\noindent
\textbf{Works on Attention Maps.}
A common assumption behind Transformers and other attention networks is that intra-sequence relationships can be automatically captured through attention mechanisms. In practice, however, it is questionable whether the attention layer learns reasonable dependencies among input tokens. Many efforts have attempted to analyze attention maps generated by attention mechanisms. Raganato \textit{et al.}~\cite{raganato2018analysis} analyze Transformer models for machine translation and show that some attention heads are able to capture certain relations implicitly: lower layers tend to learn more about grammar, while higher layers tend to encode more about semantics. Tang \textit{et al.}~\cite{tang2018self} show that Transformer models are less capable of inducing syntactic relations than their recurrent neural network counterparts.
Tay \textit{et al.}~\cite{tay2020synthesizer} argue that explicit token-token interactions are not important, and dot product attention can be replaced with synthetic attention maps. Furthermore, there is debate about whether the intermediate representation provided by the attention mechanism helps explain the predictions of the model~\cite{jain2019attention,wiegreffe2019attention}. In short, the attention maps induced by existing attention mechanisms are not good enough. Furthermore, there are some successful attempts to combine convolutional and self-attention layers to enrich image and text representations~\cite{bello2019attention, wu2020lite}. To the best of our knowledge, our work is one of the first to treat attention maps as multichannel images and utilize dedicated deep neural networks for pattern extraction and evolution. We believe this is a promising direction that deserves more research in future work.

\noindent
\textbf{Time-Series Representation.}
For time-series representation, the widely adopted network structures for encoding include GRU~\cite{guo2018multidimensional}, LSTM~\cite{malhotra2016lstm, sagheer2019unsupervised}, dilated convolutions~\cite{yu2016multi,franceschi2019unsupervised} and ResNet~\cite{wang2017time}. Among them, the most commonly used architecture is dilated convolution owing to its outstanding efficiency and generality. 
Recently, Zerveas \textit{et al.}~\cite{zerveas2021transformer} propose a transformer-based model for multivariate time-series representation, which achieves competitive results in both time-series regression and classification tasks. 
Nevertheless, the transformer-based models are still outperformed by dilated convolution networks on some datasets, especially for smaller ones where the generalization ability is very crucial. 
Attention mechanisms have also been widely applied in time-series-related tasks~\cite{qin2017dual,shih2019temporal,liang2018geoman,lyu2018improving}. For instance, GeoMAN~\cite{liang2018geoman} integrates local and global attentions to capture the complex spatial correlations between different time series. Qin \textit{et al.}~\cite{qin2017dual} utilize attentions to adaptively extract the relevant series across different time steps. SAnD~\cite{song2018attend} employs a masked self-attention mechanism to model dependencies among various sequences, achieving superior performance in clinical time-series benchmarks.
Another challenge of time-series representation is how to learn generalized representation vectors for a diverse and complex data distribution.
Most existing methods are based on some comparison functions~\cite{paparrizos2019grail} or contrasting learning~\cite{franceschi2019unsupervised,tonekaboni2021unsupervised,eldeletime2021}.
T-Loss~\cite{franceschi2019unsupervised} adopts a triplet loss on randomly cropped subseries to enhance time-series representations. TNC~\cite{tonekaboni2021unsupervised} uses a constant-length sampling window and selects adjacent segments as positive samples based on local smoothness. TS2Vec~\cite{yue2021ts2vec} proposes a contextual consistency protocol, which treats the contextual representations at the same timestamp in two masked subseries as a positive pair and leverages a hierarchical contrastive loss to capture multiscale knowledge in different abstraction levels. 

\noindent
\textbf{Computer Vision.}
Attention-based networks show the strong results in cases of general representation learning~\cite{woo2018cbam,hu2018squeeze} and relevant down-stream tasks including segmentation~\cite{fu2019dual,li2021towards}, detection~\cite{carion2020end,hu2018relation}, generation~\cite{zhang2019self}, low-level vision~\cite{dai2019second} and 3D vision~\cite{xie2018attentional,guo2021pct}. 
In particular, Bello \textit{et al.}~\cite{bello2019attention} propose Attention Augmented (AA-) ResNets by integrating the attention mechanism into ResNet. 
Several previous works~\cite{parmar2019stand,bello2020lambdanetworks,zhao2020san,involution} propose to replace the convolution with attention-like operators. Parmar \textit{et al.}~\cite{parmar2019stand} apply stand-alone self-attention layers to image classification, while LambdaNetworks~\cite{bello2020lambdanetworks} adopt lambda layers which model long-range interactions between a query and a structured set of context elements. Recently, Vision Transformer (ViT)~\cite{dosovitskiy2020image,touvron2021training,liu2021swin} divides an image into a sequence of patches as input to a standard transformer~\cite{vaswani2017attention}, and the following works~\cite{liu2021swin,Wu2021CvTIC,zhang2022eatformer,Xia_2022_CVPR,han2023reference,li2023transformer} explore the locality of self-attention. Moreover, several works~\cite{li2022video,yuan2022polyphonicformer,li2022panoptic,xu2022fashionformer,zhou2022transvod} combine the Detection Transformer~\cite{detr} for the unified down-stream visual tasks. All these methods are specifically designed for image tasks and do not generalize to other tasks in NLP or time-series domain.

\noindent
\textbf{Natural Language Processing.}
Transformer~\cite{vaswani2017attention} is solely composed of self-attention and feed-forward layers. It is more parallelizable than Recurrent Neural Networks (RNNs) and demonstrates superiority in large-scale natural language models. The pre-trained text representation models, BERT~\cite{devlin2018bert} and RoBERTa~\cite{liu2019roberta}, are based on deep bidirectional Transformers. After being pre-trained on a large-scale language corpus with Masked Language Model (MLM) and Next Sentence Prediction (NSP) losses, the BERT-style models can be fine-tuned with just one additional output layer to create state-of-the-art performance for a wide range of text-oriented applications. The pre-trained text generation models, such as T5~\cite{raffel2019exploring} and GPT-3~\cite{brown2020language}, utilize a left-to-right Transformer architecture to predict the text sequence token by token. UniLM~\cite{dong2019unified} further enables a joint optimization of natural language understanding (NLU) and natural language generation (NLG) tasks in a unified framework.
There are other research directions, including relative positional representations~\cite{shaw2018self}, long sequence understanding~\cite{kitaev2020reformer,zhang2021poolingformer,sukhbaatar2019adaptive}, tree-based transformer~\cite{shiv2019novel}, and AutoML-based evolved transformer~\cite{so2019evolved}. Our proposal puts emphasis on the evolution of attention maps, which is orthogonal to these works.

\section{Convolution-enhanced Evolving Attention Networks}


In a vanilla attention network, attention maps are generated by each attention layer separately without explicit interactions between different abstraction levels. However, as we have argued in the introduction, an independent attention layer does not learn the underlying token dependencies as cross-layer transferable knowledge, leading to unsatisfactory generality capability. Therefore, we adopt a residual convolutional module that generalizes attention maps in the current layer based on the inherited knowledge from previous layers. We name attention networks augmented with the proposed mechanism as Convolution-enhanced Evolving Attention Networks and add a prefix ``EA-" in front of different backbone models.

In the following content, we first introduce the generic evolving attention mechanism over given attention maps in Section \ref{sec:evolving_attention} and the instantiation of Evolving Attention Transformer in Section \ref{sec:ea_transformer}. Then, we propose Evolving Attention Dilated Convolutional Transformer (EA-DC-Transformer) for time-series representation tasks in Section \ref{sec:ea_dc_transformer}. Finally, the extension to other attention networks will be discussed in Section \ref{sec:other_attention_networks}.

\subsection{The Evolving Attention Mechanism}
\label{sec:evolving_attention}
The representation of a text or time-series sequence can be written as $\mathbf{X} \in \mathbb{R}^{N \times C}$, where $N$ denotes the sequence length and $C$ is the dimension size. For an image representation, the conventional shape is $(H, W, C)$, where $H, W$ and $C$ denote height, width and channel size of the image, respectively. In order to apply a standard attention network to the image representation, we flatten its shape as $\mathbf{X} \in \mathbb{R}^{N \times C}$, where $N=HW$, while each pixel serves as an individual token in the attention network.



In ordinary attention networks, the attention maps of each layer are computed independently without explicit interaction with each other. In contrast, in EA-enhanced Attention Neural Networks, we establish explicit skip connections between adjacent attention maps.
Assuming there are $K$ heads, we have $K$ output attention maps for each layer. We represent them together by the tensor $A \in \mathbb{R}^{N \times N \times K}$ ($N$ is the sequence length), which can be thought of as a $N \times N$ image with $K$ channels. Taking this as input, we employ a 2D convolutional layer with a $3 \times 3$ kernel to generalize the attention map. The output channel is also set to $K$ so that attention maps for all heads can be jointly generated. We apply ReLU activations after each 2D convolutional layer to provide non-linearity and sparsity for the model.
Finally, the result attention map is combined with input attention map and fed into the Softmax activation layer. Mathematically, 
\begin{equation}
\begin{aligned}
\label{eq:combine_ratio}
    & \mathbf{A}^{i}_{input} = \alpha \cdot \mathbf{A}^{i-1}_{logit} + (1 - \alpha) \cdot \text{Attention}(\mathbf{X})^{i}_{logit}, \\
    & \mathbf{A}^{i}_{logit} = \beta \cdot \text{Conv}(\mathbf{A}^{i}_{input}) + (1 - \beta) \cdot \mathbf{A}^{i}_{input}, \\
    & \mathbf{A}^{i} = \text{softmax}(\mathbf{A}^{i}_{logit}),
\end{aligned}
\end{equation}
where $\mathbf{A}^{i-1}_{logit}$ is the attention logit matrix of the previous block; $\text{Attention}(\mathbf{X})^{i}_{logit}$ is the logit matrix calculated by the current attention layer (before Softmax); $\mathbf{A}^{i}_{input}$ is the combined matrix after residual connection, as the input of the convolution module. $\text{Conv}(\cdot)$ represents a 2D-convolutional layer with ReLU activations. $\alpha, \beta \in [0, 1]$ are the hyper-parameters of linear combination. In our experiments, the values of $\alpha$ and $\beta$ are chosen empirically on the validation set for each task. 



\begin{figure}[t]
	\centering
        \includegraphics[width=\linewidth]{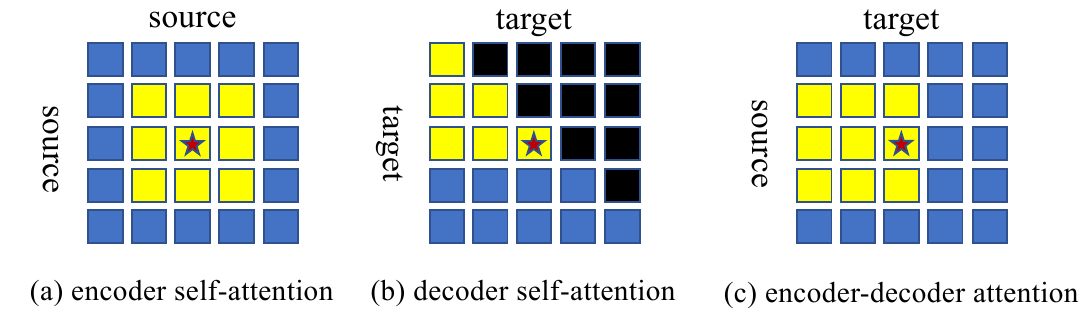}
         \caption{Different convolution masks for three kinds of attentions.}
         \label{fig:text}
\end{figure}

\textbf{Convolution Masks for Seq2Seq Models.}
In a sequence-to-sequence attention network, there are three kinds of attention, namely, encoder self-attention, decoder self-attention, and encoder-decoder attention. For the encoder, we employ standard convolutions that take into account all surrounding pixels within the local window. For the decoder, we employ a convolutional mask to prevent envisioning subsequent tokens. Figure \ref{fig:text} visualizes three convolutional masking strategies for attention maps, respectively, where the current token is highlighted by a star. In the figure, yellow pixels are considered for convolution, while other pixels are masked.

Figure \ref{fig:text}(a) shows the the operation for encoder attention maps, which is a standard convolution. Instead, the decoder only takes upper-left pixels in the self-attention maps for convolution. As illustrated by Figure \ref{fig:text}(b), the upper-right pixels (black color) are permanently masked, and other pixels (blue color) are not calculated for the current token. This forms a convolution kernel with a receptive field of 6, which can be achieved in two steps: (1) performing a standard $3 \times 3$ convolution with a mask in the upper-right corner; (2) after convolution, shifts the entire attention matrix 1 pixel down and 1 pixel to the right. 
As shown in Figure \ref{fig:text}(c), the encoder-decoder attention only pays attention to the left pixels in the convolution kernel to prevent information leakage. This can be achieved with a standard $3 \times 3$ convolution, where 1 pixel is shifted to the right.

\subsection{Evolving Attention Transformer}
\label{sec:ea_transformer}

Specifically, the transformer architecture augmented by evolving attention is illustrated in Figure \ref{fig:overview}(b). Each EA-Transformer block consists of four modules, including \textit{Attention Map Generation}, \textit{Convolution-based Evolving Attention}, \textit{Value Projection}, and \textit{Feed-Forward Layers}. 
The residual connections between attention maps (highlighted by the red lines) are designed to facilitate the information flow of attention patterns with regularization effects. Note that we omit layer norms in the figure for brevity. The convolution-based evolving attention module has been described in Section \ref{sec:evolving_attention}. Other components will be presented in the following sub-sections sequentially. 

\textbf{Self-Attention Layer.}
Given the input representation $\textbf{X}$, the attention maps can be calculated as follows. First, we compute the query and key matrices for each attention head through linear projections, \textit{i.e.}, $\mathbf{Q} = \mathbf{XW}^Q, \mathbf{K} = \mathbf{XW}^K$, where $\mathbf{Q}$ and $\mathbf{K}$ denote query and key matrices respectively, $\mathbf{W}^Q$ and $\mathbf{W}^K$ are linear projection parameters. 
Then, the attention map is derived by a scaled dot-product operation: 
\begin{equation}
\begin{aligned}
    \mathbf{A} = & \text{Attention}(\mathbf{X}) = \text{softmax}(\frac{\mathbf{Q}\mathbf{K}^\top}{\sqrt{d}} + \mathbf{R}). 
\label{dot_product_attention}
\end{aligned}
\end{equation}
Here $\mathbf{A}$ denotes the attention map, $d$ is the hidden dimension size.
To inject sequential information into the model, we incorporate positional encoding to the input representation. The positional encoding can be either absolute or relative, and we follow the original implementation for each baseline model.
The absolute positional embedding~\cite{vaswani2017attention} is added to token embedding $\mathbf{X}$ directly. 

For relative positional representation~\cite{shaw2018self}, a matrix $\mathbf{R}=\{\mathbf{r}_{ij}\}$ is added to reflect relative positional encoding. For text data, we have $\mathbf{r}_{ij} = \mathbf{q}_i^T\mathbf{e}_{i-j}$, where $\mathbf{e}_{i-j}$ is a trainable embedding vector in terms of the relative indices for two tokens. For image data, we adopt two separate embedding vectors for height and width~\cite{bello2019attention}.
\begin{equation}
\mathbf{r}_{ij} = \mathbf{q}_i^\top \mathbf{e}^H_{h(j)-h(i)} + \mathbf{q}_i^\top \mathbf{e}^W_{w(j)-w(i)},
\end{equation}
where $\mathbf{q}_i$ is the query representation for the $i^{th}$ pixel, $\mathbf{e}^H$ and $\mathbf{e}^W$ represent for trainable embedding vectors of height and width, respectively, $h(i)$ and $h(j)$ are the height indices for the $i^{th}$ and $j^{th}$ pixels respectively, and $w(\cdot)$ denotes the index of width.

\textbf{Value Projection and Feed-Forward Layers.}
Given the original attention map $\mathbf{A}^{i}_{input}$, we generate a new attention map $\mathbf{A}$ through the generic evolving attention mechanism described in Section \ref{sec:evolving_attention}. The rest of an EA-Transformer block includes the same \textit{value projection} and \textit{position-wise feed-forward} layers as the standard transformer block. 
The value projection layer can be expressed as: 
\begin{equation}
\begin{aligned}
    \mathbf{H}_k = \mathbf{A}_k\mathbf{X}\mathbf{W}_k^V,   ~~~ \mathbf{H} = (\mathbf{H}_{1} \oplus \mathbf{H}_{2} \oplus ... \oplus \mathbf{H}_{K})\mathbf{W}^O, \\
\end{aligned}
\end{equation}
where $\mathbf{A}_k$ is the attention map of the $k^{th}$ head, $\mathbf{W}_k^V$ is the parameter of value projection, and $\mathbf{H}_k$ is the result representation after value projection.
After that, the representations of all heads are concatenated (denoted by $\oplus$) and fed into a linear projection layer with parameter $\mathbf{W}^O$. 
Finally, the block ends with two position-wise feed-forward layers:
\begin{equation}
    \text{EA-Transformer}(\mathbf{X}) = \text{ReLU}(\mathbf{H}\mathbf{W}_1 + \mathbf{b}_1)\mathbf{W}_2 + \mathbf{b}_2.
\end{equation}
By convention, $\textbf{W}_1$ has four times the dimensions of $\mathbf{W}^O$ and $\mathbf{W}_2$, forming a bottleneck structure. For brevity, we omit the layer norms and skip connections here.

\subsection{Evolving Attention-enhanced Dilated Convolutional Transformer for Time-Series Representation}
\label{sec:ea_dc_transformer}
Dilated convolutions~\cite{yu2016multi,van2016wavenet} and transformers~\cite{vaswani2017attention,zerveas2021transformer} are most common network architectures used in the time-series domain, which achieve state-of-the-art performance in time-series classification and regression benchmarks. While the transformer captures sparse connectivity patterns in time series data, dilated convolution has a global receptive field with plausible local inductive bias, which benefits both efficiency and accuracy.
To combine the advantages of two kinds of architectures, we integrate dilated convolutions into evolving attention-enhanced transformer networks by concatenating their representations. We name this new model as Evolving Attention-enhanced Dilated Convolutional (EA-DC-) Transformer. 

\subsubsection{Network Architecture}
The network architecture with two adjacent EA-DC-Transformer blocks is illustrated in Figure \ref{fig:overview}(c). The 1D-convolution module denotes a multi-layer dilated convolution~\cite{franceschi2019unsupervised}, where the first layer has a dilation $s=1$; the second layer has a dilation $s=2$, and the $n$-th layer has a dilation $s=2^n$. Besides the hyper-parameters $\alpha$ and $\beta$ which control the importance of residual links and convolution-based attention evolution modules separately, we also introduce a hyper-parameter $p$ to balance the representation dimensions from transformer and dilated convolution. The operations of the $i$-th EA-DC-Transformer layer are defined in the following equations. 
\begin{equation}
\begin{aligned}
\label{eq:ea-dc}
    & \mathbf{A}^{i}_{input} = \alpha \cdot \mathbf{A}^{i-1}_{logit} + (1 - \alpha) \cdot \text{Attention}(\mathbf{X})^{i}_{logit}, \\
    & \mathbf{A}^{i}_{logit} = \beta \cdot \text{CNN}(\mathbf{A}^{i}_{input}) + (1 - \beta) \cdot \mathbf{A}^{i}_{input}, \\
    & \mathbf{A}^{i} = \text{softmax}(\mathbf{A}^{i}_{logit}), ~~\mathbf{H}_k^{i} = \mathbf{A}_k^{i}\mathbf{X}\mathbf{W}_k^V,  \\
    & \mathbf{H}^{i} = (\mathbf{H}^{i}_{1} \oplus \mathbf{H}^{i}_{2} \oplus ... \oplus \mathbf{H}^{i}_{K})\mathbf{W}^O,~Dim(\mathbf{H}^{i}) = p \cdot d,\\
    & \mathbf{D}^{i} = \text{DilatedConv}(H^{i-1}),~Dim(\mathbf{D}^{i}) = (1-p) \cdot d, \\
    & \mathbf{R}^{i} = \text{Concat}(\mathbf{H}^{i}, \mathbf{D}^{i}), ~~Dim(\mathbf{R}^{i}) = d, \\
    & \text{EA-DC-T}(\mathbf{X})^{i} = \text{ReLU}(\mathbf{R}^{i}\mathbf{W}_1 + \mathbf{b}_1)\mathbf{W}_2 + \mathbf{b}_2,
\end{aligned}
\end{equation}
where $d$ stands for the hidden dimension size, other notations have similar meanings to Equation (1)-(6). Note that when $p=0$, the model degenerates to a Dilated Convolution, while when $p=1$, it is identical to an EA-Transformer architecture.
Empirically, we allocate 25\% dimension to the transformer branch and the rest 75\% to dilated convolution. We find that this setting generally works well for most datasets, and a sensitivity analysis of value $p$ will be given in Section \ref{sec:hyper-parameter}.

\subsubsection{Optimization}
\textbf{Pre-training.}
Given a time series input $\mathbf{X} \in \mathbb{R}^{T \times C}$, where $T$ is the number of timestamps, and $C$ is the multivariate dimension; we randomly mask a portion $r$ of the input values. We define the mask as $\mathbf{M} = \{m_{t,i} | t \in T, i \in C\}$, where $m_{t, i} \in \{0, 1\}$, and the set of masked indices is $I = \{(t, i)~|~m_{t,i} = 0\}$. First, we estimate the masked values by a reconstruction function on top of the learned representation output from EA-DC-transformer. 
\begin{equation}
\begin{aligned}
\label{eq:mask_z}
    & \hat{\mathbf{Z}} = \text{EA-DC-T}(\mathbf{X \cdot M}), ~~\hat{\mathbf{X}} = \mathbf{W}_{rc} \hat{\mathbf{Z}} + \mathbf{b}_{rc},
\end{aligned}
\end{equation}
where $\mathbf{W}_{rc}$ and $\mathbf{b}_{rc}$ are learnable parameters for value reconstruction.
Next, an Mean Squared Error (MSE) loss is used for each masked dimension and timestamp pair $(t, i)$, forcing the reconstruction values to be consistent to the original input values. 
\begin{equation}
\label{eq:mse}
     \mathcal{L}_{pretrain} = \frac{1}{|I|}\sum_{(t, i) \in I} (\hat{\mathbf{X}}_{t,i} - \mathbf{X}_{t,i})^2. \\
\end{equation}
Following \cite{zerveas2021transformer}, we mask 15\% of input values, \textit{i.e.}, $r=0.15$. 

\noindent \textbf{Fine-tuning.}
After the unsupervised pre-training stage, the model is fine-tuned on each downstream task, allowing update of all model parameters.  
The representation for each time-series input can be derived by $\mathbf{Z} = \text{EA-DC-T}(\mathbf{X})$. Note that the representation $\mathbf{Z}$ is calculated without masking on the original input $\mathbf{X}$. 
We consider two kinds of downstream tasks in our experiments. 
For time-series regression tasks, we apply linear regression to the learned representation, and utilize an MSE loss between the estimated and ground truth values:
\begin{equation}
     \hat{y} = \mathbf{W}_{lr} \mathbf{Z} + \mathbf{b}_{lr}, ~~\mathcal{L}_{lr} = (\hat{y} - y)^2. \\
\end{equation}
For time-series classification tasks, an MLP classifier is stacked upon the learned representation, while a cross-entropy loss function is adopted for training:
\begin{equation}
\hat{\mathbf{y}} = \text{Softmax}(\mathbf{W}_{cl} \mathbf{Z} + \mathbf{b}_{cl}), ~~\mathcal{L}_{cl} = \mathbf{y} \log(\hat{\mathbf{y}}).
\end{equation}
Here $\mathbf{y} \in \mathbb{R}^{n}$ and $n$ denotes the number of classes. 

\subsection{Extension to other Attention Networks}
\label{sec:other_attention_networks}
Convolution-enhanced evolving attention networks embody many kinds of attention-enhanced architectures. Here we present some other kinds of backbone attention networks, which are commonly used in natural language and computer vision fields.

\noindent \textbf{BERT-Style Models.} State-of-the-art language models, BERT~\cite{devlin2018bert}, and its extensions, such as RoBERTa~\cite{liu2019roberta} and T5~\cite{raffel2019exploring}, adopt transformer architectures and can be pre-trained by self-supervised losses such as Masked Language Model (MLM) and Next Sentence Prediction (NSP). For these types of models, we employ the same network architecture as an Evolving Attention Transformer. The pre-trained parameters of BERT are uploaded as initialization, while the additional parameters in the convolution-based evolution mechanism are randomly initialized and updated with the original BERT parameters in the continuous pre-training or fine-tuning stages.

\noindent \textbf{Attention Augmented ResNet.}
Attention Augmented (AA-) ResNet~\cite{bello2019attention} demonstrated that traditional CNN models~\cite{lecun2010convolutional} can benefit from attention mechanisms in computer vision. It concatenates the image representations computed by self-attention and convolutional neural networks to capture both local and global information. 
Our EA-AA-ResNet architecture enhances its attention mechanism by bridging the attention maps from different layers and extracting common attention patterns through a convolution module. 
In the image representation layer, half of the dimension size is generated by convolution, and the other half is reserved for self-attention.

\noindent \textbf{Vision Transformer.}
Vision Transformer (ViT) is inspired by the Transformer scaling successes in NLP, which adopts a standard Transformer architecture with only a few modifications for images, that is, splitting an image into patches and utilizing the sequential embedding vectors of these patches as input in the same way as natural language tokens. Thus, it is straightforward to take ViT as a special kind of attention neural network and enhance it by evolving attention techniques. Vision Transformer (ViT) obtains excellent results when pre-trained at sufficient scale and have good transfer ability to the downstream tasks.

\noindent \textbf{Graph Attention Networks.}
Graph Attention Networks (GATs)~\cite{velivckovic2018graph,wang2019heterogeneous} operate on graph-structured data through masked self-attention layers to address the shortcomings of graph convolutions~\cite{defferrard2016convolutional,kipf2017semi}. The advantages of GATs lie in their efficiency of parallel computation, while they do not require a prior knowledge for the underlining graph structure. The evolving attention mechanism can be applied to GATs directly with $1 \times 1$ convolution kernels, keeping equivalent to the permutations of tokens. A neighborhood ranking method is necessary when using a kernel size larger than one. More details will be discussed in Section \ref{sec:limitation}.

\noindent \textbf{Other Attention Types.} Our evolving attention mechanism is generic and not restricted to the dot-product attention (Equation (\ref{dot_product_attention})). There are other attention types, such as multiplicative, additive and concatenative attentions, which are formulated in the equations below.
\begin{equation}
\begin{aligned}
    & \mathbf{A}_m = \text{softmax}(\mathbf{Q^\top}\mathbf{W}_1\mathbf{K}), \\
    & \mathbf{A}_a = \text{softmax}(\mathbf{W}_2^\top \text{tanh}(\mathbf{Q^\top}(\mathbf{W}_3\mathbf{K} + \mathbf{W}_4\mathbf{Q}), \\
    & \mathbf{A}_c = \text{softmax}(\mathbf{W}_5^\top \text{tanh}(\mathbf{Q^\top}\mathbf{W}_6\mathbf{[K;Q]}),
\label{attention_types}
\end{aligned}
\end{equation}
where $\mathbf{A}_m$, $\mathbf{A}_a$ and $\mathbf{A}_c$ stand for multiplicative, additive and concatenative attention maps respectively; $\{\mathbf{W}_i | i = 1,2,..,6\}$ are trainable parameters in the attention functions. For all these attention types, we can easily make an extension to the evolving attention mechanism, \textit{i.e.}, applying residual convolutions between the corresponding attention maps in adjacent layers. 
\begin{table*}[t]
\centering
\caption{Comparison of RMSE/\#Parameters for different models on time-series regression datasets }
\scalebox{1.0}{
    \renewcommand\arraystretch{1.0}
    \centering
    \begin{tabular}{lccccccc}
    \toprule
     \textbf{Dataset} & \textbf{LSTM} & \textbf{GRU} & \textbf{ResNet} & \textbf{Dilated Conv} & \textbf{Transformer} & \textbf{DC-T (ours)} & \textbf{EA-DC-T (ours)}  \\ 
     \midrule
   AppliancesEnergy	&	3.844	/	1.2 	M	&	4.151	/	0.9 	M	&	3.369	/	0.6 	M	&	3.711	/	0.6 	M	&	3.663	/	0.6 	M	&	3.035	/	0.8 	M	& \textbf{	2.957}	/	0.8 	M \\
    BenzeneConcentr	&	7.936	/	1.2 	M	&	6.919	/	0.9 	M	&	2.889	/	0.5 	M	&	2.758	/	0.5 	M	&	1.576	/	0.5 	M	&	1.127	/	0.5 	M	& \textbf{	0.758}	/	0.5 	M	\\
    BeijingPM10	&	101.863	/	0.3 	M	&	101.452	/	0.2 	M	&	95.22	/	0.1 	M	&	96.927	/	0.1 	M	&	98.035	/	0.2 	M	&	91.993	/	0.2 	M	& \textbf{	91.774}	/	0.2 	M	 \\
    BeijingPM25	&	64.715	/	0.3 	M	&	65.667	/	0.2 	M	&	64.54	/	0.1 	M	&	64.813	/	0.1 	M	&	64.874	/	0.4 	M	&	59.425	/	0.5 	M	& \textbf{	59.118}	/	0.5 	M	 \\
    LiveFuelMoisture	&	43.316	/	0.3 	M	&	44.19	/	0.3 	M	&	44.723	/	0.2 	M	&	43.457	/	0.2 	M	&	44.874	/	0.2 	M	&	43.326	/	0.3 	M	& \textbf{	43.261}	/	0.3 	M	 \\
    IEEEPPG	&	34.814	/	1.4 	M	&	26.961	/	1.1 	M	&	46.593	/	2.7 	M	&	39.633	/	2.7 	M	&	33.848	/	5.8 	M	&	30.075	/	5.8 	M	& \textbf{	23.14}	/	5.8 	M	 \\

     \midrule
     \textbf{Avg Rank} & 5.2	&	5.5	&	4.5	&	4.7	&	4.8	&	2.3	& \textbf{	1.0	} \\
     \textbf{Avg Rel. diff.} & 0.251	&	0.182	&	0.035	&	0.009	&	\text{-}0.072	&	\text{-}0.173	& \textbf{\text{-}0.231} \\

     \bottomrule
    \end{tabular}
    }
    \label{ts_reg}
\end{table*}

\begin{table*}[t]
\centering
\caption{Comparison of accuracy/\#Parameters for different models on time-series classification datasets }
\scalebox{1.0}{
    \renewcommand\arraystretch{1.0}
    \centering
    \begin{tabular}{lccccccc}
    \toprule
     \textbf{Dataset} & \textbf{LSTM} & \textbf{GRU} & \textbf{ResNet} & \textbf{Dilated Conv} & \textbf{Transformer} & \textbf{DC-T (ours)} & \textbf{EA-DC-T (ours)}  \\ 
     \midrule
        EthanolConcentration	&	0.224	/	0.8 	M	&	0.251	/	0.8 	M	&	0.27	/	0.7 	M	&	0.272	/	0.7 	M	&	0.272	/	0.6 	M	&	0.293	/	0.8 	M	& \textbf{	0.307}	/	0.8 	M	 \\
        FaceDetection	&	0.644	/	0.3 	M	&	0.661	/	0.3 	M	&	0.667	/	0.5 	M	&	0.669	/	0.5 	M	&	0.664	/	0.4 	M	&	0.672	/	0.5 	M	& \textbf{	0.685}	/	0.5 	M	\\
        Handwriting	&	0.156	/	0.5 	M	&	0.159	/	0.5 	M	&	0.25	/	0.4 	M	&	0.251	/	0.4 	M	&	0.282	/	0.7 	M	&	0.319	/	0.7 	M	& \textbf{	0.329}	/	0.7 	M	\\
        Heartbeat	&	0.686	/	0.4 	M	&	0.675	/	0.3 	M	&	0.704	/	0.2 	M	&	0.698	/	0.2 	M	&	0.699	/	0.3 	M	&	0.715	/	0.3 	M	& \textbf{	0.724} /	0.3 	M	 \\
        JapaneseVowels	&	0.827	/	0.3 	M	&	0.871	/	0.2 	M	&	0.954	/	0.4 	M	&	0.955	/	0.4 	M	&	0.979	/	0.4 	M	&	0.982	/	0.4 	M	& \textbf{	0.985}	/	0.5 	M	\\
        PEMS-SF	&	0.742	/	0.7 	M	&	0.798	/	0.5 	M	&	0.811	/	0.5 	M	&	0.803	/	0.5 	M	&	0.724	/	0.5 	M	&	0.825	/	0.5 	M	& \textbf{	0.829}	/	0.5 	M	 \\
        SelfRegulationSCP1	&	0.836	/	1.5 	M	&	0.877	/	1.2 	M	&	0.887	/	0.9 	M	&	0.882	/	0.9 	M	&	0.885	/	0.7 	M	&	0.893	/	0.8 	M	& \textbf{	0.896}	/	0.8 	M	 \\
        SelfRegulationSCP2	&	0.459	/	1.6 	M	&	0.511	/	1.3 	M	&	0.481	/	1.0 	M	&	0.454	/	1.0 	M	&	0.491	/	0.8 	M	&	0.493	/	0.9 	M	& \textbf{	0.526}	/	0.9 	M	 \\
        SpokenArabicDigit	&	0.98	/	0.4 	M	&	0.985	/	0.3 	M	&	0.979	/	0.2 	M	&	0.984	/	0.2 	M	&	0.984	/	0.2 	M	&	0.987	/	0.3 	M	& \textbf{	0.991}	/	0.3 	M	 \\
        UWaveGestureLibrary	&	0.435	/	1.5 	M	&	0.774	/	1.2 	M	&	0.843	/	1.7 	M	&	0.846	/	1.7 	M	&	0.852	/	1.9 	M	&	0.861	/	2.0 	M	& \textbf{	0.873}	/	2.0 	M	 \\
        InsectWingbeat	&	0.575	/	0.3 	M	&	0.574	/	0.3 	M	&	0.610	/	0.5 	M	&	0.619	/	0.5 	M	&	\textbf{0.639}	/	0.4 	M	&	0.630	/	0.5 	M	& 0.622	/	0.5 	M \\

    \midrule
    \textbf{Avg Rank} &	6.5	&	5.5	&	4.5	&	4.5	&	3.8	&	2.1	& \textbf{	1.2	} \\
    \textbf{Avg Accuracy} & 0.597	&	0.649	&	0.678	&	0.676	&	0.679	&	0.697	& \textbf{	0.706	} \\

     \bottomrule
    \end{tabular}
    }
    \label{ts_cla}
\end{table*}

\section{Experiments}
To thoroughly prove the effectiveness and fundamental contributions of the novel evolving attention module, we conduct extensive experiments on a variety of tasks that widely cover time-series representation, natural language understanding, machine translation and image classification. For each task, we enhance the state-of-the-art attention network in that domain with the evolving attention mechanism to obtain the evolving attention-enhanced variant and add an ``EA-" prefix before the model name, e.g., EA-BERT for BERT. We not only compare the performance with strong baseline models, but also conduct ablation studies to demonstrate the effectiveness of the evolving attention mechanism.

\subsection{Time-Series Representation}
\label{exp:ts}
As introduced in the methodology, we propose the \textit{DC-Transformer} architecture that takes an integration of dilated convolutions into the transformer network as the backbone model. As we will show, this novel architecture already achieves state-of-the-art performance at various time-series representation tasks. Moreover, the \textit{EA-DC-Transformer} equipped with evolving attention mechanism further shows superiority to the strong backbone model.

\noindent \textbf{Settings.}
EA-DC-Transformer consists of $n$ blocks, each with $2$ evolving attention-enhanced convolution layers. The kernel size of each layer defaults to 3. 
We adopt Rectified Adam optimizer~\cite{liu2019variance} with $\beta_1 = 0.9$ and $\beta_2 = 0.99$. The dropout ratio is 0.1, and the learning rate is set to 1e-3 empirically. Hyper-parameters are tuned in the following search space on validation set: hidden dimension \{64, 128\}, $\alpha=$\{0.1, 0.3, 0.5, 0.7, 0.9\} and $\beta=$\{0.1, 0.3, 0.5, 0.7, 0.9\}. 
The dimension ratio to combine convolution and transformer branches is chosen from $p=$\{0.125, 0.25, 0.375, 0.5, 0.625, 0.75, 0.875\}. 
The number of blocks $n$ is chosen from \{2, 3, 4, 5\}.

The architecture of transformer and its hyper-parameters follow the paper of Time-Series Transformer (TST)~\cite{zerveas2021transformer}. Both of LSTM~\cite{hochreiter1997long} and GRU~\cite{cho2014learning} encoders adopt a standard implementation in Pytorch~\cite{paszke2019pytorch}. Dilated Convolution follows the implementation of \cite{franceschi2019unsupervised} and ResNet follows \cite{he2016deep}. The kernel size of convolution layer is 3. For all baseline models, we adopt the same optimizer, dropout ratio and learning rate as those in DC-Transformer and EA-DC-Transformer. Hidden dimension and number of blocks are optimized in the same search space.

Following previous works~\cite{zerveas2021transformer,yue2021ts2vec,franceschi2019unsupervised}, we evaluate the model quality for time-series representation on two tasks, \textit{i.e.}, time-series regression and time-series classification, which predict a corresponding regression value or target category based on each time-series input, respectively.
We leverage a representative range of 6 datasets from the Monash Time Series Regression Archive~\cite{tan2020monash} to ensure the diversity with respect to dimensionality and length of time-series samples. 
Both Root Mean Squared error (RMSE) and average relative difference are taken as evaluation metrics. The ``average relative difference" metric $\mathbf{r}_{j}$ for each model $j$ over $N$ datasets and $M$ compared models can be computed by:
\begin{equation}
\begin{aligned}
\label{self-attention}
    \mathbf{r}_{j} = \frac{1}{N} \sum_{i=1}^N\frac{\mathbf{r}_{ij}-\overline{\mathbf{R}_{i}}}{\overline{\mathbf{R}_{i}}},\qquad \overline{\mathbf{R}_{i}} = \frac{1}{M} \sum_{k=1}^M \mathbf{r}_{ik} \\
\end{aligned}
\end{equation} 
where $\mathbf{r}_{ij}$ is the RMSE of model $j$ on dataset $i$ and $M$ is the number of compared models.
For classification tasks, we include 11 multivariate datasets from the UEA Time Series Classification Archive~\cite{bagnall2018uea} with diverse characteristics in terms of the number of classes, dimensionality and length of time series samples. We report accuracy, average accuracy, and average rank among all compared models.

\noindent \textbf{Results.}
It can be seen from Table \ref{ts_reg} that the proposed backbone model DC-Transformer generally outperforms other baselines, ranking at 2.3 on average for time-series regression tasks. Compared to the best baseline (ResNet, average rank $4.5$), the average RMSE of DC-Transformer is reduced significantly by 20.1\%. By further utilizing the evolving attention mechanism, EA-DC-Transformer boosts the average RMSE by 9.88\% and performs the best on all regression datasets.
Also, DC-Transformer and EA-DC-Transformer generally improves the performance of existing models in time-series classification tasks.
As illustrated in Table \ref{ts_cla}, EA-DC-Transformer performs the best on 10 out of 11 datasets, achieving an average rank of $1.2$, which beats DC-Transformer (average rank $2.1$) and the best baseline model (Transformer, average rank $3.8$) by a significant margin. 
These results indicate that EA-DC-Transformer learns superior time-series representations for both regression and classification tasks. The evolving attention mechanism, which ameliorates attention maps through convolution and residual connections layer by layer, learns more precise inter-timestamp dependency patterns in multiple abstraction levels. We also report the number of parameters for all compared models in Table \ref{ts_reg} and Table \ref{ts_cla}. To summarize, the parameters of EA-DC-Transformers are on par with DC-Transformers and Transformers, while being only a slightly larger than GRUs and ResNets. 



\noindent \textbf{Ablation Study.}
We perform ablation studies on EA-DC-Transformer on four regression datasets and four classification datasets. The results are listed in Table \ref{ts_reg_ablation} and \ref{ts_cla_ablation}, respectively. According to the results, both dilated convolution structures and residual connections are essential to time-series representation learning. Replacing $3 \times 3$ kernels with $1 \times 1$ ones leads to performance degradation in most datasets, but the best performance is achieved in the dataset \textit{Bene}. We suspect it is caused by a specific distribution or characteristics of time series in this dataset.

\begin{table}
\centering
\caption{Ablation study for time-series regression}
\scalebox{1.0}{
    \renewcommand\arraystretch{1.0}
    \begin{tabular}{lcccc}
    \toprule
     \textbf{Model} & \textbf{Appli.} & \textbf{Bene.} & \textbf{Bei10.} & \textbf{Bei25.} \\ 
     \midrule
     Transformer-Base & 3.663 & 1.576 & 98.035 & 64.874\\
     EA-DC-Transformer-Base & \textbf{2.957} & 0.758 & \textbf{91.774} & \textbf{59.118}\\
     $~$ \textit{w/o Convolution} & 3.256 & 0.801 & 92.343 & 61.243\\
     $~$ \textit{w/o Skip Connection} & 3.258 & 0.937 & 92.353 & 60.290\\
     $~$ \textit{with $1 \times 1$ Convolution} & 3.197 & \textbf{0.736} & 94.566 & 61.820\\
     $~$ \textit{with $5 \times 5$ Convolution} & 3.147 & 0.865 & 95.163 & 60.805\\
     \bottomrule
    \end{tabular}
}
    \label{ts_reg_ablation}
\end{table}

\begin{table}
\centering
\caption{Ablation study for time-series classification}
\scalebox{1.0}{
    \renewcommand\arraystretch{1.0}
    \begin{tabular}{lcccc}
    \toprule
     \textbf{Model} & \textbf{UWave.} & \textbf{Face.} & \textbf{Hand.} & \textbf{Heart.} \\ 
     \midrule
     Transformer-Base & 0.852 & 0.664 & 0.251 & 0.698\\
     EA-DC-Transformer-Base & \textbf{0.873} & \textbf{0.685} & \textbf{0.329} & \textbf{0.724}\\
     $~$ \textit{w/o Convolution} & 0.872 & 0.667 & 0.323 & 0.696 \\
     $~$ \textit{w/o Skip Connection} &0.869 & 0.682 & 0.324 & 0.698 \\
     $~$ \textit{with $1 \times 1$ Convolution} & 0.866 & 0.672 & 0.303 & 0.704 \\
     $~$ \textit{with $5 \times 5$ Convolution} & 0.859 & 0.678 & 0.342 & 0.715 \\
     \bottomrule
    \end{tabular}
}
    \label{ts_cla_ablation}
\end{table}

\subsection{Natural Language Understanding}
\label{exp:bert}

\begin{table*}[t]
\centering
    \caption{Comparison of different models on GLUE benchmark for text understanding. The metrics for SST-2, MNLI, QNLI and RTE are accuracy. The metrics for MRPC and QQP are F1/accuracy. The Matthew's correlation coefficient and Pearson/Spearman correlation coefficient are used for CoLA and STS-B, respectively. The average GLUE score is calculated by the first metric for each dataset.} 
    \scalebox{0.95}{
    \renewcommand\arraystretch{1.1}
    \centering
    \begin{tabular}{lccccccccccc}
    \toprule
     \textbf{Model} & \textbf{\#Params} & \textbf{\#FLOPs} & \textbf{Avg} & \textbf{CoLA} & \textbf{SST-2} & \textbf{MRPC} & \textbf{STS-B} & \textbf{QQP} & \textbf{MNLI-m/-mm} & \textbf{QNLI} & \textbf{RTE} \\ 
     \midrule
     BERT-Base (dev) & 109.5M & 6.3G & 83.1 & 56.3 & 92.7 & 89.9/85.6 & 89.5/89.7 & 91.3/88.3 & 84.5/84.5 & 91.3 & 69.3 \\
     \textbf{EA-BERT-Base (dev)} & 110.4M & 6.8G & \textbf{84.4} & \textbf{61.6} & \textbf{93.6} & \textbf{90.9/87.0} & \textbf{90.0/90.2} & \textbf{91.5/88.5} & \textbf{84.9/85.0} & \textbf{92.0} & \textbf{70.8} \\
     \midrule
     BERT-Base (test) & 109.5M & 6.3G & 79.3 & 52.6 & 93.5 & 87.5/82.7 & 84.9/83.4 & \textbf{71.3/89.4} & 84.6/83.8 & \textbf{91.0} & 68.9 \\
     \textbf{EA-BERT-Base (test)} & 110.4M & 6.8G & \textbf{80.3} & \textbf{56.7} & \textbf{93.8} & \textbf{88.1/84.1} & \textbf{86.3/85.3} & \textbf{71.3}/89.3 & \textbf{85.0/84.3} & \textbf{91.0} & \textbf{69.9} \\
     \midrule
     T5-Base (dev) & 220.2M & 9.1G & 83.5 & 53.1 & 92.2 & 92.0/88.7 & 89.1/88.9 & 88.2/91.2 & 84.7/85.0 & 91.7 & 76.9 \\
     Synthesizer (dev) & 272.0M & 11.3G & 84.1 & 53.3 & 92.2 & 91.2/87.7 & 89.3/88.9 & 88.6/91.4 & 85.0/84.6 & \textbf{92.3} & 81.2 \\
     \textbf{EA-T5-Base (dev)} & 221.2M & 9.9G & \textbf{84.6} & \textbf{53.7} & \textbf{93.1} & \textbf{92.3/89.0} & \textbf{89.6/89.1} & \textbf{88.8/91.9} & \textbf{85.1/85.0} & \textbf{92.3} & \textbf{81.5} \\
     \midrule
     BERT-Large (dev) & 335.0M & 12.2G & 84.3 & 60.5 & 94.9 & 89.3/85.4 & 87.6/86.5 & 92.1/89.3 & 86.8/85.9 & 92.7 & 70.1 \\
     RealFormer (dev) & 335.0M & 12.2G & 84.8 & 59.8 & 94.0 & \textbf{90.9}/87.0 & \textbf{90.1/89.9} & 91.3/88.3 & 86.3/86.3 & 91.9 & \textbf{73.7}\\
     \textbf{EA-BERT-Large (dev)} & 336.7M & 12.9G & \textbf{85.7} & \textbf{62.9} & \textbf{95.2} & \textbf{90.9}/\textbf{89.4} & 89.7/88.2 & \textbf{92.4/90.1} & \textbf{87.9/86.8} & \textbf{93.9} & 72.4 \\
     \midrule
     RoBERTa-Large (dev) & 355.0M & 12.7G & 88.5 & 65.8 & 95.9 & 91.2/87.8 & 92.1/92.0 & 92.2/89.6 & 90.2/90.1 & 94.9 & 80.7 \\
     \textbf{EA-RoBERTa-Large (dev)} & 356.7M & 13.3G & \textbf{89.4} & \textbf{68.1} & \textbf{96.8} & \textbf{92.0/88.7} & \textbf{92.3/92.2} & \textbf{92.3/89.7} & \textbf{90.6/90.4} & \textbf{95.0} & \textbf{85.3}\\
     \midrule
     RoBERTa-Large (test) & 355.0M & 12.7G & 86.3 & 61.2 & 96.2 & 91.4/88.5 & 91.2/90.5 & 73.3/89.6 & 89.8/89.4 & 94.8 & 80.0 \\
     \textbf{EA-RoBERTa-Large (test)} & 356.7M & 13.3G & \textbf{87.3} & \textbf{66.1} & \textbf{96.4} & \textbf{92.1/89.2} & \textbf{91.5/90.7} & \textbf{73.4/89.8} & \textbf{90.5/89.8} & \textbf{94.9} & \textbf{83.7} \\
     \bottomrule
    \end{tabular}
    }
    \label{tab:glue}
\end{table*}

Pre-trained Language Models (PLMs) like BERT become popular in recent years. 
These models are based on bi-directional transformer architectures and pre-trained by a large corpus.
The evolving attention mechanism can be easily plugged into an existing checkpoint of vanilla PLMs and achieve significant improvement through continuous training. To demonstrate this advantage, we choose GLUE benchmark~\cite{wang2018glue} for an empirical study.

\noindent \textbf{Settings.}
BERT's encoder network consists of multiple transformer blocks. In EA-BERT, we replace each transformer block with EA-Transformer shown in Figure \ref{fig:overview}(b). 
We separately load the pretrained checkpoints for BERT-Base, T5-Base, BERT-Large, and RoBERTa-Large and fine-tune them individually for each downstream task on task-specific training data.
The additional parameters introduced by the evolving attention mechanism are randomly initialized and jointly trained with other parameters.
We use Adam optimizer~\cite{kingma2014adam} with epsilon 1e-8.
The dropout ratio is set as 0.1 empirically. The hyper-parameters are tuned in the following search spaces on the validation set: learning rate \{1e-4, 1e-5, 2e-5\}, batch size \{8, 16\}, number of training epochs \{2, 3, 5\}, $\alpha=$ \{0.1, 0.2, 0.4\} and $\beta=$ \{0.1, 0.2, 0.4\}. 



\noindent \textbf{Results.}
The comparison between BERT-style models is shown in Table \ref{tab:glue}. 
The T5-Base and BERT-Large models are evaluated on the development set to be comparable with existing baselines. Other models are evaluated on both dev and test sets. The test set results are obtained through the GLUE online evaluation portal\footnote{\textit{https://gluebenchmark.com/}}.
In different downstream tasks, EA-BERT generally outperforms vanilla BERT. Specifically, EA-BERT-Base, EA-T5-Base, EA-BERT-Large and EA-RoBERTa-Large have average scores of 84.4, 84.6, 85.7, and 89.4 on the GLUE benchmark, achieving +1.3, +1.1, +1.4, and +0.9 absolute improvements, compared to corresponding backbones on the development set, respectively. On the test set, the average scores are consistently lifted by 1.0 points on top of BERT-Base and RoBERTa-Large models. All improvements are achieved by loading existing checkpoints and fine-tuning parameters with limited training time. This is also an attractive advantage for large-scale applications.
Furthermore, EA-BERT significantly improves the performance of CoLA, showing its excellent generalization ability on small datasets.
Moreover, the EA-enhanced architectures demonstrate more benefits than two recent works, i.e., Realformer~\cite{he2020realformer} and Synthesizer~\cite{tay2020synthesizer}. Realformer utilizes residual connections over attention maps but does not exploit convolution-based evolution modules. Synthesizer synthesizes a separate attention map and mixes it with vanilla self-attention.

\begin{table}
\centering
\caption{Ablation study for text understanding}
\scalebox{0.95}{
    \renewcommand\arraystretch{1.1}
    \begin{tabular}{lcccc}
    \toprule
     \textbf{Model} & \textbf{CoLA} & \textbf{SST-2} & \textbf{MRPC} & \textbf{MNLI} \\ 
     \midrule
     BERT-Base & 58.4 & 92.9 & 89.9/85.6 & 84.7/84.5\\
     EA-BERT-Base & \textbf{61.6} & \textbf{93.6} & \textbf{90.9/87.0} & \textbf{84.9/85.0}\\
     $~$ \textit{w/o Convolution} & 52.1 & 93.4 & 89.6/84.8 & 84.5/84.4\\
     $~$ \textit{w/o Skip Connection} & 52.9 & 93.5 & 90.2/85.8 & 84.2/83.8\\
     $~$ \textit{with $1 \times 1$ Convolution} & 57.5 & 93.5 & 89.8/85.5 & 84.7/84.8\\
     $~$ \textit{with $5 \times 5$ Convolution} & 59.2 & 93.5 & 90.6/86.5 & 84.8/85.0\\
     \bottomrule
    \end{tabular}
}
    \label{tab:tu_ablation}
\end{table}

\noindent \textbf{Ablation Study}
We conduct ablation studies on EA-BERT-Base on four text understanding datasets with different data scales. According to the results in Table \ref{tab:tu_ablation}, the advantage of EA-BERT comes from both convolution-based pattern extraction and residual connections. Removing the convolutional module or replacing it with a $1 \times 1$ kernel results in a significant performance drop. Skip connection is also advantageous. Without it, we would lose a lot of gain.
Meanwhile, the $5 \times 5$ kernel also produces competitive results, but with more parameters. 

\begin{table*}[t]
\centering
\caption{BLUE scores on machine translation datasets}
\scalebox{1.0}{
    \renewcommand\arraystretch{1.1}
    \centering
    \begin{tabular}{lccccc}
    \toprule
     \textbf{Model} & \textbf{\#Params} & \textbf{\#FLOPs (En-De)} & \textbf{IWSLT’14 De-En} & \textbf{WMT'14 En-De} & \textbf{WMT'14 En-Fr} \\ 
     \midrule
     Transformer-Lite & 2.48M & 158.31M & 33.32 & 21.11 & 33.22 \\
     \textbf{EA-Transformer-Lite} & 2.49M & 163.46M & \textbf{33.80} & \textbf{21.63} & \textbf{34.12} \\
     \midrule
     Transformer-Base & 44.14M & 2.68G & 34.55 & 27.47  & 40.79 \\
     \textbf{EA-Transformer-Base} & 44.15M & 2.70G & \textbf{35.30}  & \textbf{27.56} & \textbf{41.54} \\
     \bottomrule
    \end{tabular}
    }
    \label{tab:mt}
\end{table*}

\begin{table}[t]
\centering
\caption{Ablation study for machine translation}
\scalebox{1.0}{
    \renewcommand\arraystretch{1.1}
    \begin{tabular}{lcccc}
    \toprule
     \textbf{Model} & \textbf{De-En} & \textbf{En-De} & \textbf{En-Fr} \\ 
     \midrule
     Transformer-Lite & 33.21 & 21.11 & 33.22 \\
     EA-Transformer-Lite & \textbf{33.80} & \textbf{21.63} & \textbf{34.12} \\
     $~~$ \textit{w/o Encoder Convolution} & 32.84 & 21.07 & 33.51 \\
     $~~$ \textit{w/o Decoder Convolution} & 33.56 & 21.45 & 33.72 \\
     $~~$ \textit{w/o Encoder-Decoder Convolution} & 33.59 & 21.41 & 33.73 \\ 
     $~~$ \textit{w/o Skip Connection} & 33.70 &  21.43 & 34.06 \\
     $~~$ \textit{with $1 \times 1$ Convolution} & 33.15 & 21.20 & 33.52 \\
     $~~$ \textit{with $5 \times 5$ Convolution} & 33.54 & 21.40 & 34.08 \\
     \bottomrule
    \end{tabular}
}
 \label{mt_ablation}
\end{table}

\subsection{Machine Translation}
\label{exp:mt}
Machine translation is a common testbed for evaluating sequence-to-sequence architectures for natural language generation (NLG) tasks. In our experiments, we employ three machine translation datasets for evaluation: IWSLT'14 German-English (De-En)~\cite{cettolo2014report}, WMT'14 English to German (En-De) and WMT'14 English to French (En-Fr)~\cite{bojar2014findings}. We apply Byte Pair Encoding (BPE)~\cite{sennrich2015neural} to the source language corpus and the target language corpus, respectively.

\noindent \textbf{Settings.}
In EA-Transformer, convolution modules are applied to encoder self-attention, decoder self-attention, and encoder-decoder attention, respectively. Skip connections are only used in encoder networks as they hurt the performance of the decoder. We set $\alpha=0.1$ and $\beta=0.1$ for EA-Transformer-Lite and $\alpha=0.5$ and $\beta=0.1$ for EA-Transformer-Base.
We employ the Adam optimizer ~\cite{kingma2014adam} with $\beta_1 = 0.9$, $\beta_2 = 0.98$, and an inverse square root learning rate schedule with a linear warm-up. The warmup step is set to 4000, and the label smoothness is set to 0.1. For each task, we choose the learning rate from \{1e-4, 5e-4, 1e-3\} and the dropout ratio from \{0.1, 0.2, 0.3\} based on the validation set.
We follow the code of Hardware-Aware Transformers~\cite{wang2020hat} to report the number of parameters and FLOPs. The number of parameters is computed without word embeddings and the final Softmax, so it is on par for different datasets. We estimate the number of FLOPs on pseudo-samples of length 30 for both source and target tokens.


\noindent \textbf{Results.}
Table \ref{tab:mt} compares Transformer and EA-Transformer with different model capacities. Transformer-Lite~\cite{wu2020lite} is a light architecture with all dimensions set to 160. Transformer-Base follows the configuration in \cite{vaswani2017attention} with 6 layers for the encoder and 6 layers for the decoder. It has 8 heads, 512 dimensions for normal layers, and 2048 dimension for the first FFN layer, forming a bottleneck structure. 
As shown in Table \ref{tab:mt}, the EA-based models achieve consistent improvements across multiple datasets and network architectures while requiring only a few additional parameters and computations.


\noindent \textbf{Ablation Study.}
Ablation results for EA-Transformer-Lite are listed in Table \ref{mt_ablation}. First, we remove the convolutional modules of the encoder, decoder, and encoder-decoder attention networks, respectively. According to the results, the convolutional module is extremely important for the encoder network. It also has positive effects on decoder self-attention and encoder-decoder attention layers. Consistent with the previous conclusion, replacing the $3\times 3$ convolution with the $1\times 1$ kernel results in a large drop in BLUE score. Therefore, local inductive bias is crucial for Transformer models to evolve towards better attention patterns.


\subsection{Image Classification}
\label{exp:ic}

\begin{table}[t]
\centering
\caption{Accuracy comparison of ResNets and EA-ResNets on ImageNet classification}
\scalebox{0.95}{
    \renewcommand\arraystretch{1.1}
    \centering
    \begin{tabular}{lcccccc}
    \toprule
     \textbf{Model} & \textbf{\#Params} & \textbf{\#FLOPs} & \textbf{Top-1} & \textbf{Top-5}\\ 
     \midrule
     ResNet-34 & 21.8M & 7.4G & 73.79 & 91.43 \\
     AA-ResNet-34 & 20.7M & 7.1G & 74.33 & 91.92 \\
     \textbf{EA-AA-ResNet-34} & 20.7M& 7.9G & \textbf{75.23} & \textbf{92.35} \\
     \midrule
     ResNet-50 & 25.6M & 8.2G & 76.44 & 93.19 \\
     AA-ResNet-50 & 25.8M & 8.3G & 77.70 & 93.80 \\
     \textbf{EA-AA-ResNet-50} & 25.8M & 8.7G & \textbf{78.22} & \textbf{94.21} \\
     \midrule
     ResNet-101 & 44.5M & 15.6G & 77.87 & 93.89 \\
     AA-ResNet-101 & 45.4M & 16.1G & 78.50 & 94.02 \\
     \textbf{EA-AA-ResNet-101} & 45.4M & 17.2G & \textbf{79.29} & \textbf{94.81} \\
     \midrule
     ResNet-152 & 60.2M&  23.0G & 78.40 & 94.20 \\
     AA-ResNet-152 & 61.6M & 23.8G & 79.10 & 94.60 \\
     \textbf{EA-AA-ResNet-152} & 61.6M & 25.7G & \textbf{79.63} & \textbf{94.85} \\
     \bottomrule
    \end{tabular}
    }
    \label{tab:imagenet}
\end{table}

\begin{table}[t]
\centering
\caption{Accuracy comparison of ViTs and EA-ViTs on ImageNet classification}
\scalebox{0.95}{
    \renewcommand\arraystretch{1.1}
    \centering
    \begin{tabular}{lcccccc}
    \toprule
     \textbf{Model} & \textbf{\#Params} & \textbf{\#FLOPs} & \textbf{Top-1} & \textbf{Top-5}\\ 
     \midrule
     ViT/B-32 & 88.3M & 13.1G &  81.22& 95.82 \\
     \textbf{EA-ViT/B-32} & 88.3M & 13.4G &  \textbf{81.48}& \textbf{95.95} \\
     \midrule
     ViT/B-16 & 86.9M & 55.5G & 84.13 & 97.11 \\
     \textbf{EA-ViT/B-16} & 86.9M & 60.3G &  \textbf{84.47} & \textbf{97.22} \\
     \midrule
     ViT/L-32 & 306.6M & 45.3G &  81.12& 95.88 \\
     \textbf{EA-ViT/L-32} & 306.7M & 46.4G & \textbf{81.50} & \textbf{96.10}  \\
     \midrule
     ViT/L-16 & 304.7M & 191.2G &  85.01& 97.37  \\
     \textbf{EA-ViT/L-16} & 304.8M & 209.0G &  \textbf{85.17}& \textbf{97.42}  \\
     \bottomrule
    \end{tabular}
    }
    \label{tab:vit}
\end{table}

We take both AA-ResNet and ViT introduced in Section \ref{sec:other_attention_networks} as the backbone models for ImageNet classification. 


\noindent \textbf{Settings.} 
All models are trained by 1.28 million training images for 100 epochs on 8 TESLA V100 GPUs. We report the top-1 and top-5 accuracy on 50k validation images and leverage 10\% training data to choose the hyper-parameters. For AA-ResNet and ViT, We follow the experimental protocols described in ~\cite{bello2019attention} and ~\cite{dosovitskiy2020image}, respectively. We set $\alpha=0.5$ and $\beta=1.0$ for EA-AA-ResNet-34, and one can refer to the appendix for detailed hyper-parameter analysis and settings of other depths of architectures. For EA-ViT models, we set $\alpha=0.1$ and $\beta=0.3$. To optimize ResNets and EA-ResNets, we adopt SGD with momentum 0.9 as the optimizer while setting batch size to 256 and weight decay to 1e-4. We train the models with 100 epochs, increase the learning rate from 0 to 0.128 linearly in the first 5 epochs and then divide its value by 10 at epoch 30, 60, 80, and 90. For ViT and EA-ViT models, we adopt an open-source implementation~\footnote{https://github.com/jeonsworld/ViT-pytorch} of ViT and add the evolving attention mechanism to construct the corresponding EA-enhanced architectures. We leverage the checkpoints pre-trained on ImageNet-21k and fine-tune it on the 1.28 million training data. One can refer to ~\cite{dosovitskiy2020image} for the detailed protocol of pre-training. We use SGD with a momentum of 0.9 as the optimizer and train the models with 20k steps with a total batch size of 512. 
Following ~\cite{dosovitskiy2020image}, we scale the input images to the resolution of 384 $\times$ 384. We train the models without weight decay while using a dropout of 0.1 rate for regularization. For the first 500 steps, we increase the learning rate from 0 to 0.03 linearly, and after that, we adopt a cosine annealing strategy to adjust its value.

\noindent \textbf{Results.} 
As shown in Table \ref{tab:imagenet} and \ref{tab:vit}, both EA-AA-ResNets and EA-ViTs consistently outperform corresponding baselines by a significant margin. The EA-AA-ResNet models boost the top-1 accuracy by 1.21\%, 0.67\%, 0.80\%, and 0.67\% relatively on top of AA-ResNet-34, -50, -101, and -152, respectively. on ViT models, the EA-enhanced variants achieve 0.32\%, 0.40\%, 0.47\%, and 0.19\% relative improvements over EA-ViT/B-32, B-16, L-32, and L-16, respectively, where B/L stands for Base/Large and 32/16 represents the patch size. These numbers are statistically significant under 95\% confidence level. As visualized in Figure \ref{fig:imagenet}, the performance enhancement is consistent for different model capacities. Arguably, this is owing to better attention maps induced by the proposed evolving attention mechanism. We will show more analyses in Section \ref{sec:analysis_attention_maps} and case studies in Section \ref{sec:case_study}. 

\begin{table}
\centering
\caption{Ablation study for EA-AA-ResNet-34}
\scalebox{1.0}{
      \begin{tabular}{lcc}
    \toprule
     \textbf{Model} & \textbf{ImageNet Top-1} & \textbf{Top-5}\\ 
     \midrule
     AA-ResNet-34 & 74.33& 91.92 \\
     EA-AA-ResNet-34 & \textbf{75.23} & 92.35 \\
     $~~$ \textit{w/o Convolution} & 74.34& 91.98  \\
     $~~$ \textit{w/o Skip Connection} & 74.29& 91.85 \\
     $~~$ \textit{with $1 \times 1$ Convolution} & 74.99 & 92.20 \\
     $~~$ \textit{with $5 \times 5$ Convolution} & 75.12 & \textbf{92.55} \\
     \bottomrule
    \end{tabular}
    }
    \label{tab:ic_ablation}
\end{table}

\noindent \textbf{Ablation Study.} 
To understand the importance of each component, we conduct ablation experiments for the EA-AA-ResNet-34 architecture. In Table \ref{tab:ic_ablation}, \textit{w/o Convolution} means to remove the attention convolution module from EA-AA-ResNet ($\beta=0$); \textit{w/o Skip Connection} means to remove skip connections between adjacent attention maps ($\alpha = 0$); In addition, we also replace the standard $3 \times 3$ convolutions with $1 \times 1$ or $5 \times 5$ kernel. As indicated by the results, both convolutions and skip connections are critical to the final performance. Also, using $1 \times 1$ convolutions (analogies to FFN) leads to a significant accuracy drop, suggesting that a convolutional inductive bias is beneficial for capturing the evolution of attention patterns. Meanwhile, $5 \times 5$ convolutions work well too, but we prefer $3 \times 3$ convolutions because they require fewer parameters.
\section{Analysis}
\label{analysis}

\begin{figure}[t]\centering                                                         
\subfigure[$~$ Transformer $~~~~$ (AvgWCD=59.4)]{                    
\begin{minipage}[t]{0.15\textwidth}
\centering                                                          

\includegraphics[width=\textwidth]{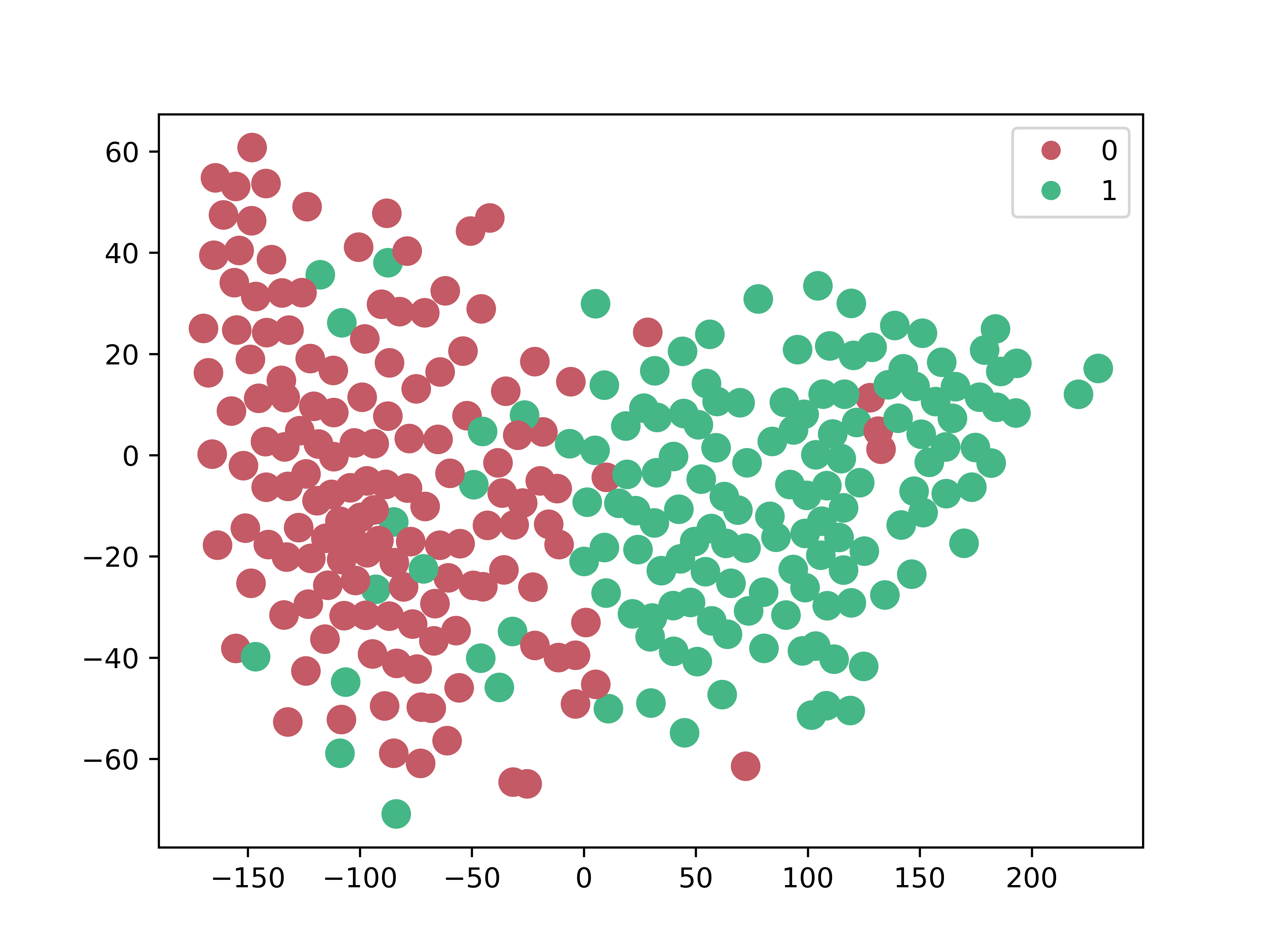}
\label{fig:side:a}
\end{minipage}}\
\subfigure[$~~$ DC-T $~~~~~~$ (AvgWCD=41.1)]{                    
\begin{minipage}[t]{0.15\textwidth}
\centering                                                          
\includegraphics[width=\textwidth]{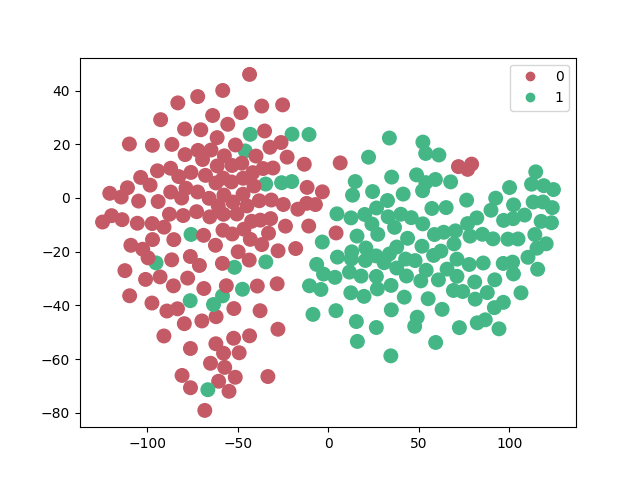}
\label{fig:side:d2}
\end{minipage}}\
\subfigure[$~~$ EA-DC-T $~~~~~$ (AvgWCD=32.8)]{                    
\begin{minipage}[t]{0.15\textwidth}
\centering                                                          
\includegraphics[width=\textwidth]{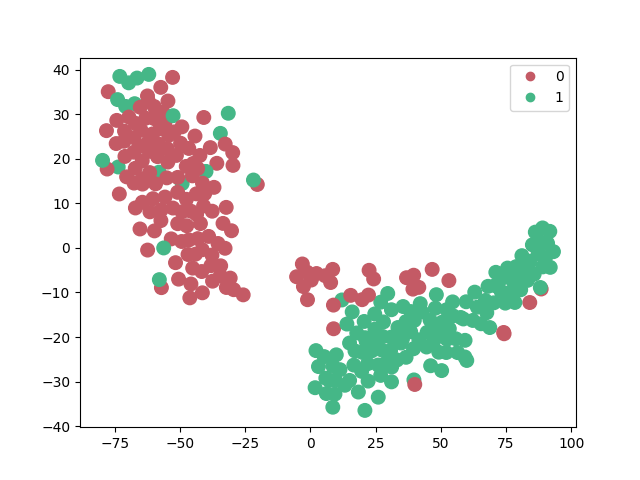}
\label{fig:side:b}
\end{minipage}}\

                                                   
\subfigure[$~$ Transformer $~~~~$ (AvgWCD=6.89)]{                    
\begin{minipage}[t]{0.15\textwidth}
\centering                                                          
\includegraphics[width=\textwidth]{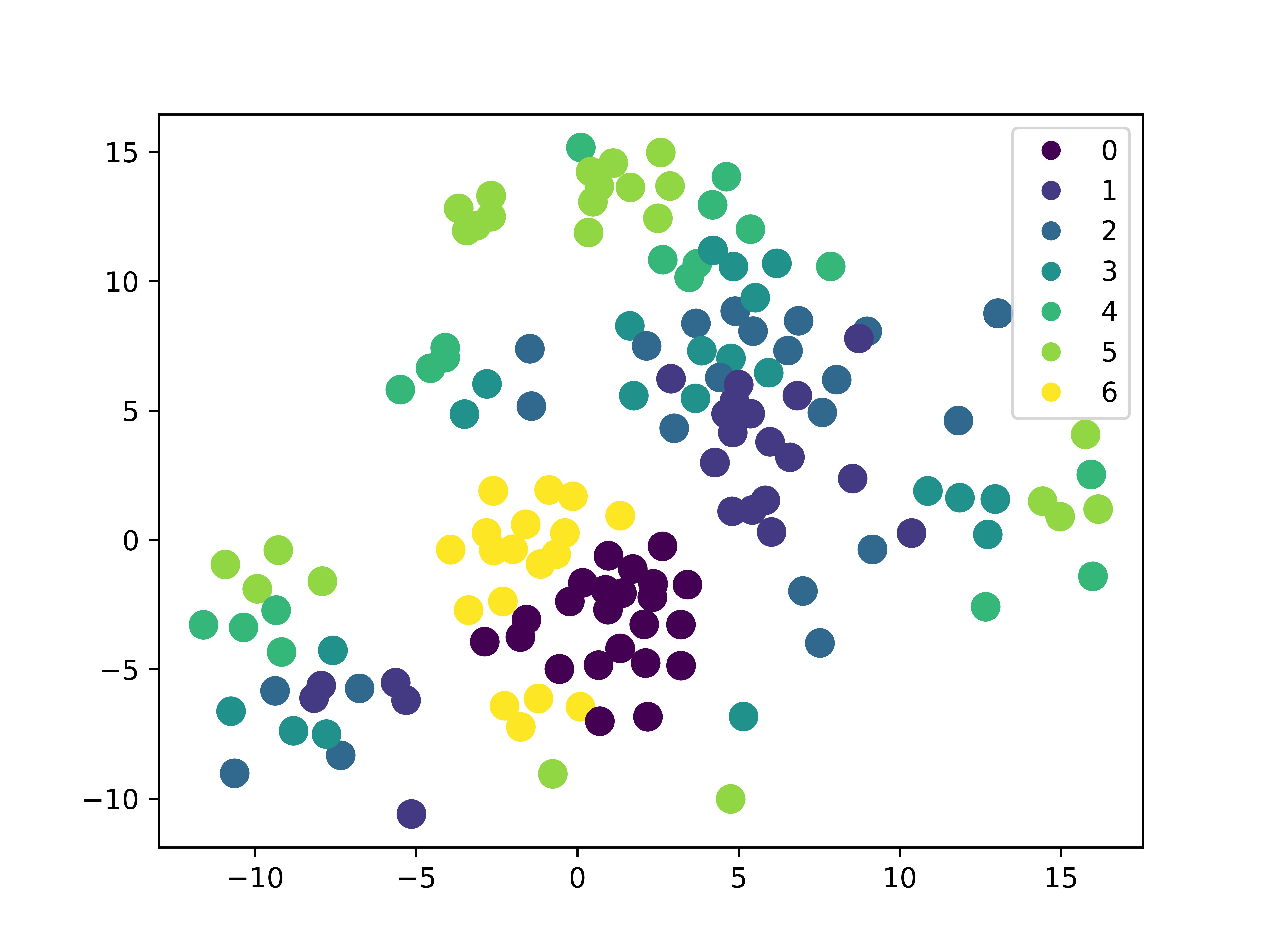}
\label{fig:side:c}
\end{minipage}}\
\subfigure[$~~$ DC-T $~~~~~~~$ (AvgWCD=4.96)]{                    
\begin{minipage}[t]{0.15\textwidth}
\centering                                                          
\includegraphics[width=\textwidth]{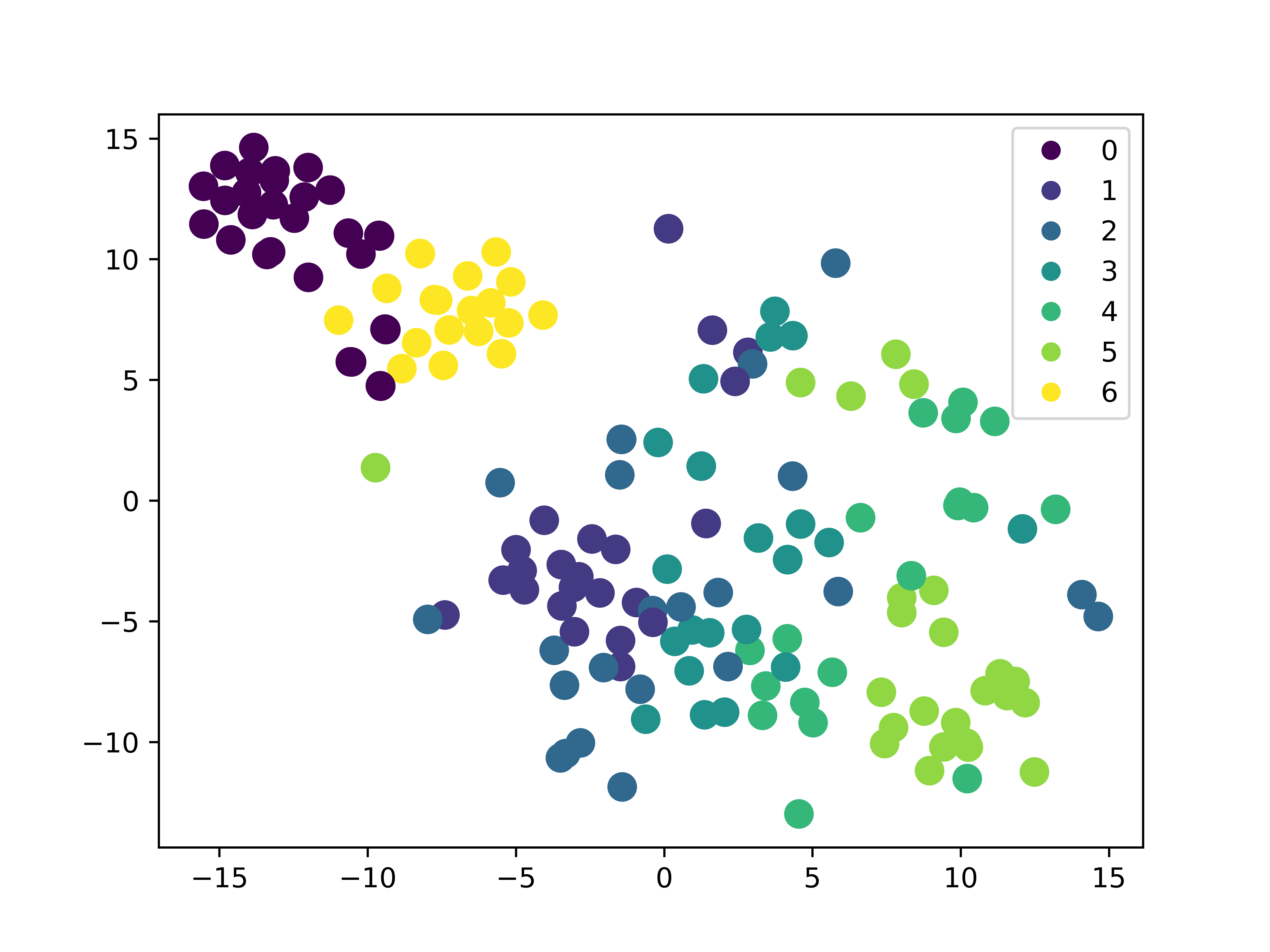}
\label{fig:side:d0}
\end{minipage}}\
\subfigure[$~~$ EA-DC-T $~~~~~$ (AvgWCD=3.86)]{                    
\begin{minipage}[t]{0.15\textwidth}
\centering                                                          
\includegraphics[width=\textwidth]{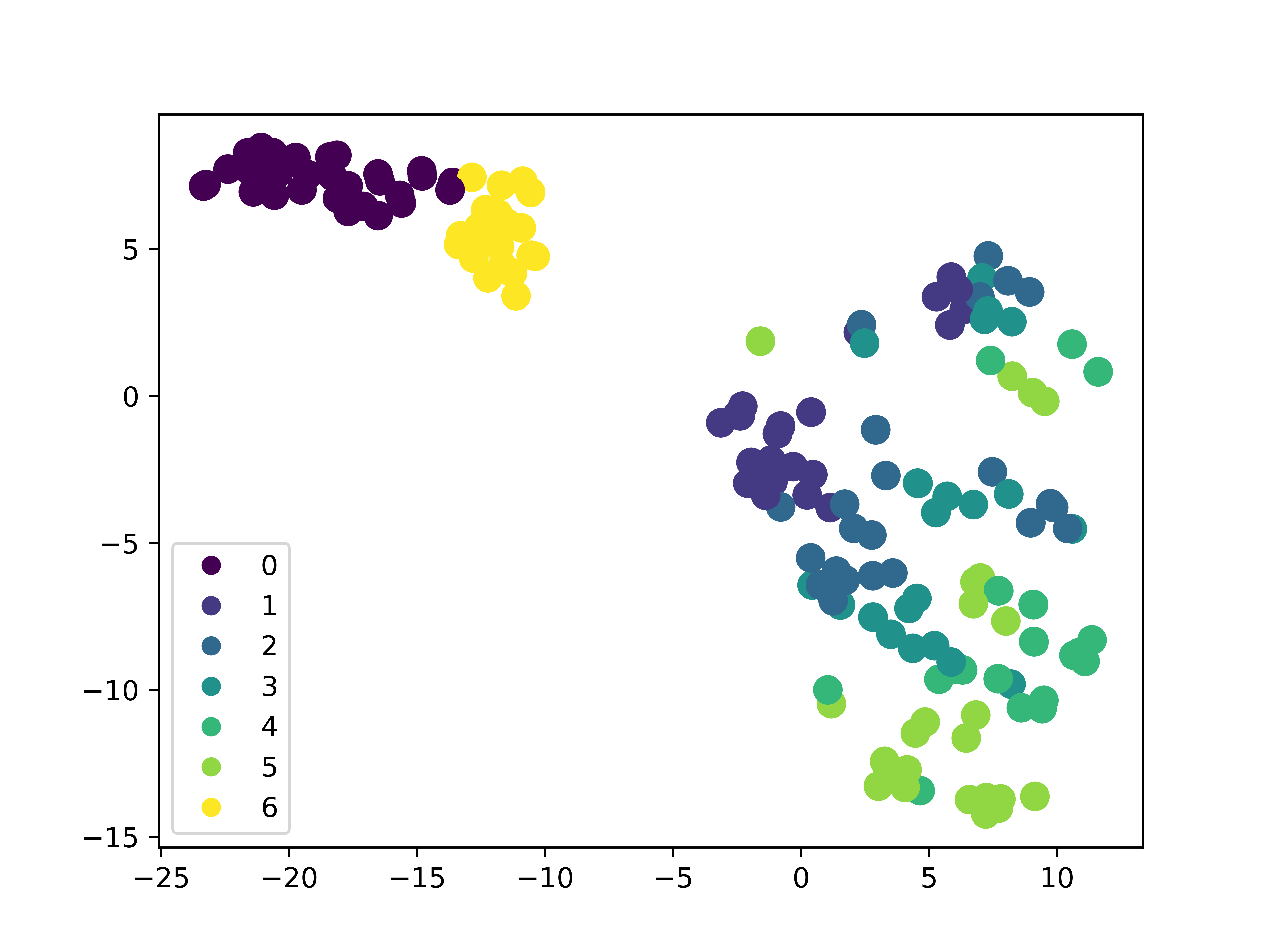}
\label{fig:side:d1}
\end{minipage}}
                                               
\subfigure[$~$ Transformer $~~~~$ (AvgWCD=6.18)]{                    
\begin{minipage}[t]{0.15\textwidth}
\centering                                                          
\includegraphics[width=\textwidth]{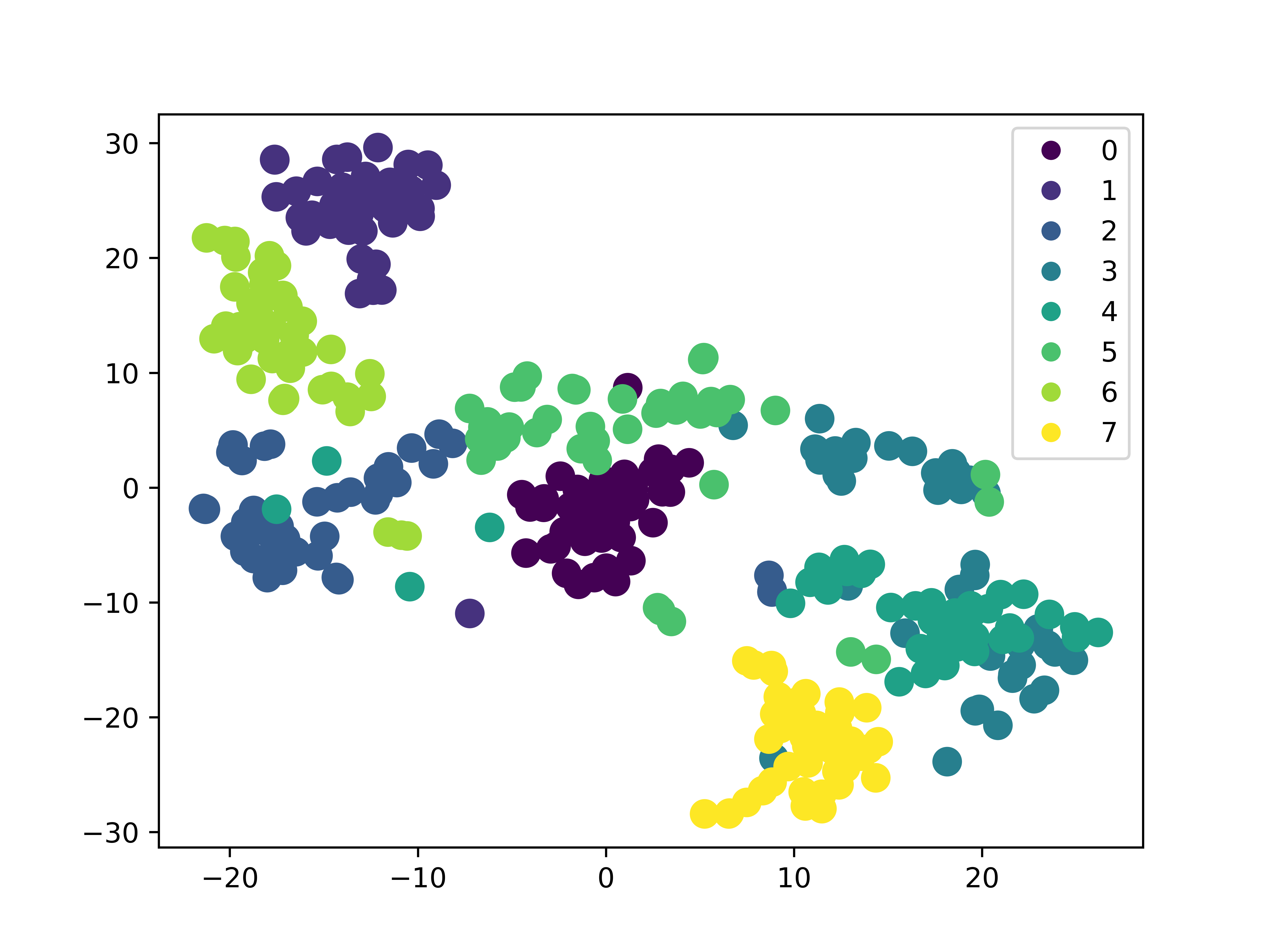}
\label{fig:side:e}
\end{minipage}}\
\subfigure[$~~$ DC-T $~~~~~~$ (AvgWCD=5.04)]{                    
\begin{minipage}[t]{0.15\textwidth}
\centering                                                          
\includegraphics[width=\textwidth]{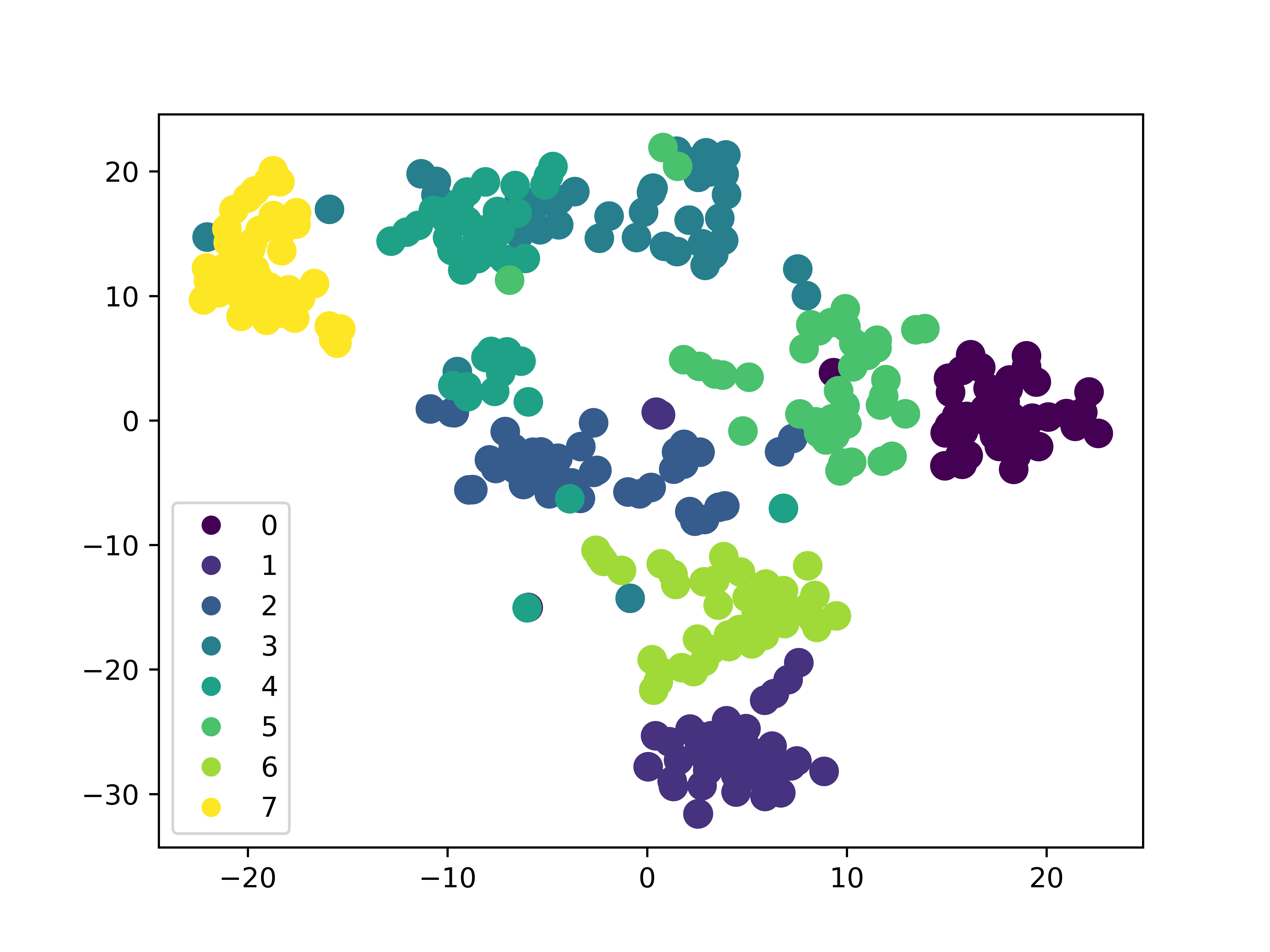}
\label{fig:side:f2}
\end{minipage}}\
\subfigure[$~~$ EA-DC-T $~~~~~$ (AvgWCD=4.96)]{                    
\begin{minipage}[t]{0.15\textwidth}
\centering                                                          
\includegraphics[width=\textwidth]{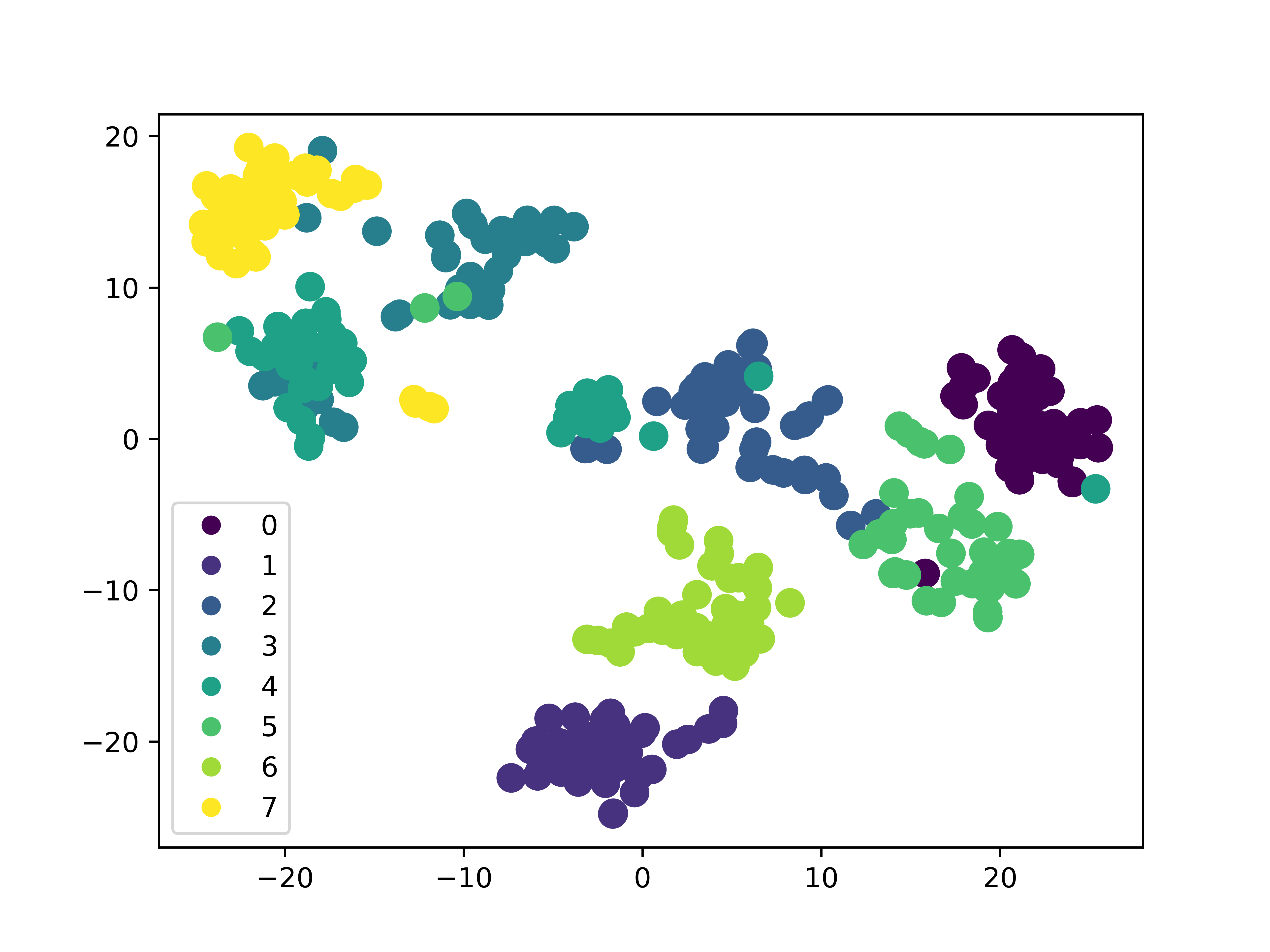}
\label{fig:side:f}
\end{minipage}}
\caption{Representations of Transformer and EA-DC-Transformer on three time-series datasets. Each row corresponds to one dataset, namely SelfRegulationSCP1, PEMS-SF and UWaveGestureLibrary, respectively from the top to bottom.} 
\label{fig:representation_compare}                                                        
\end{figure}

\subsection{Learned Representations}
We provide a detailed analysis of learned time-series representations on three time-series classification datasets, PEMS-SF, SelfRegulationSCP1 and UWaveGestureLibrary. We utilize t-SNE~\cite{van2008visualizing} to reduce the dimensions of final representations to 2 and plot them in Figure~\ref{fig:representation_compare}, wherein each color represents a specific class. 
As indicated by Figure \ref{fig:representation_compare}, Transformer fails to clearly distinguish the instances of different categories, whereas EA-DC-Transformer produces much better representations with discriminating capability.
to quantify the effectiveness of learned representations, we evaluate clustering results via \textit{average within-cluster distance}: 
\begin{equation}
\begin{aligned}
\label{awcss}
    Avg WCD = \frac{1}{N}\sum_{i=1}^k\sum_{p\in C_{i}} \sqrt{(q-m_{i})^2}, \\
\end{aligned}
\end{equation} 
where $N$ is the number of samples, $C_i$ denotes the $i$-th category for a classification task, $q$ stands for an instance in $C_i$, and $m_i$ is the center of all instance representations in $C_i$. The smaller $AvgWCD$ is, the better representation the model obtains. Compared to Transformer and DC-Transformer, EA-DC-Transformer is 36.18\% and 14.2\% better, respectively, for an average $AvgWCD$ score on three datasets.

 \begin{figure*}[t]
  \centering
        \subfigure[Input]{
        \begin{minipage}[b]{0.12\linewidth}
        \includegraphics[width=\linewidth]{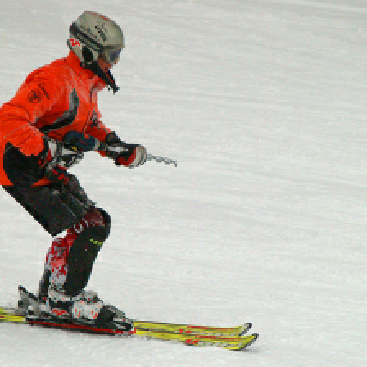}
        \includegraphics[width=\linewidth]{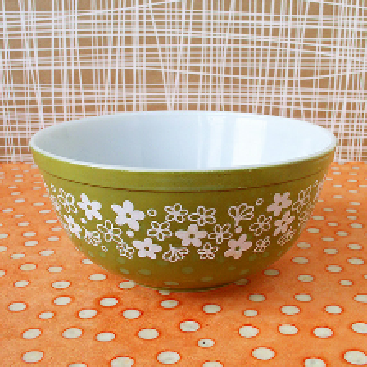}
        \includegraphics[width=\linewidth]{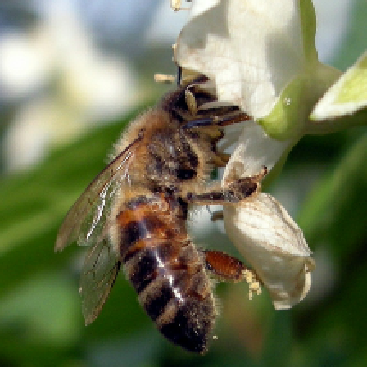}
        \end{minipage}
        }
        \subfigure[AA-16]{
        \begin{minipage}[b]{0.12\linewidth}
        \includegraphics[width=\linewidth]{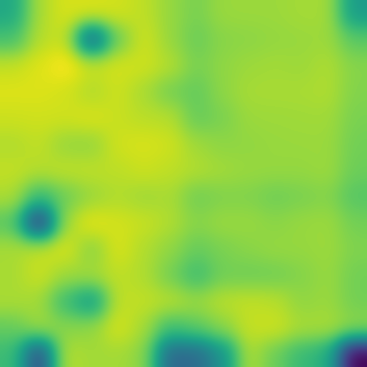}
        \includegraphics[width=\linewidth]{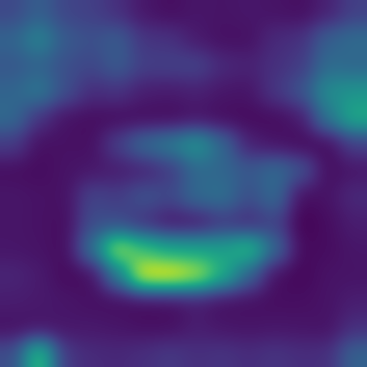}
        \includegraphics[width=\linewidth]{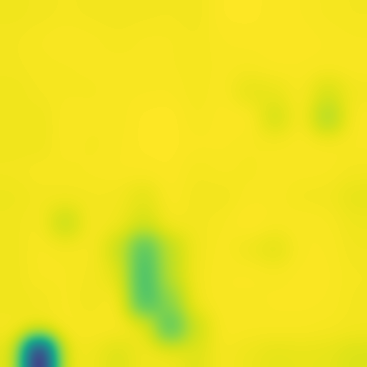}
         \end{minipage}
        }
         \subfigure[AA-17]{
        \begin{minipage}[b]{0.12\linewidth}
        \includegraphics[width=\linewidth]{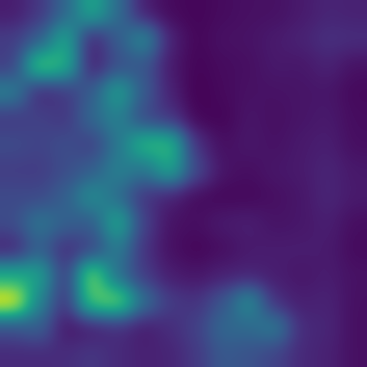}
        \includegraphics[width=\linewidth]{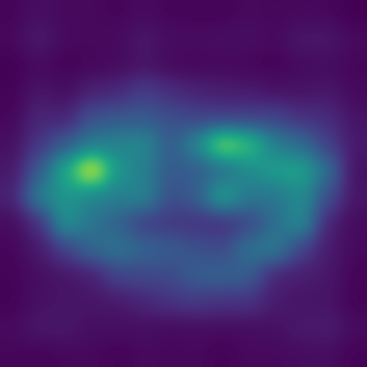}
        \includegraphics[width=\linewidth]{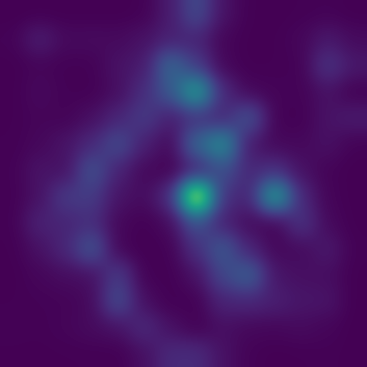}
\end{minipage}}
         \subfigure[AA-18]{
        \begin{minipage}[b]{0.12\linewidth}
        \includegraphics[width=\linewidth]{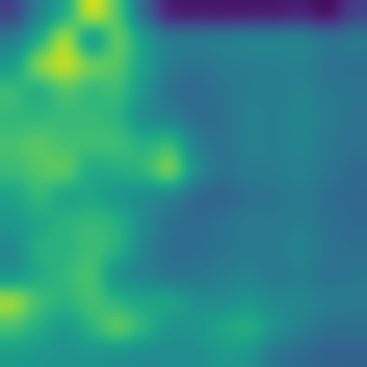}
        \includegraphics[width=\linewidth]{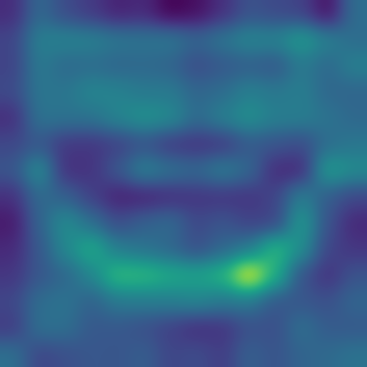}
        \includegraphics[width=\linewidth]{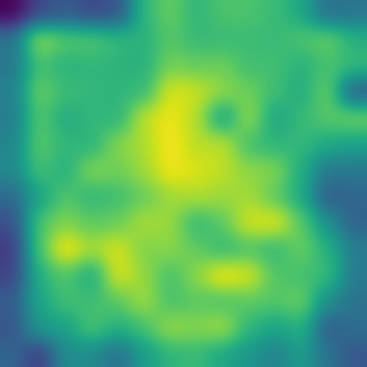}
\end{minipage}}
        \subfigure[EA-AA-16]{
        \begin{minipage}[b]{0.12\linewidth}
        \includegraphics[width=\linewidth]{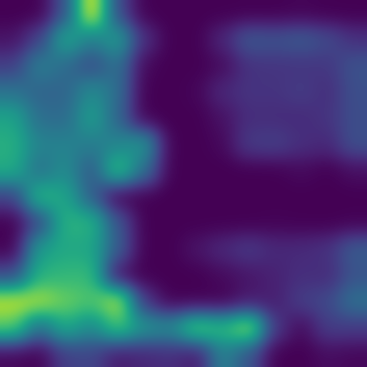}
        \includegraphics[width=\linewidth]{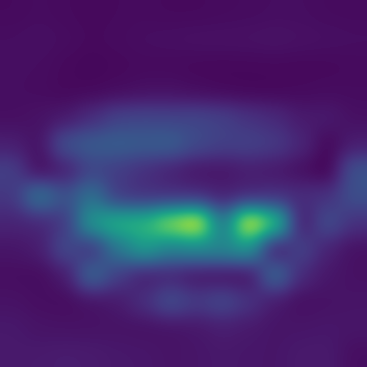}
        \includegraphics[width=\linewidth]{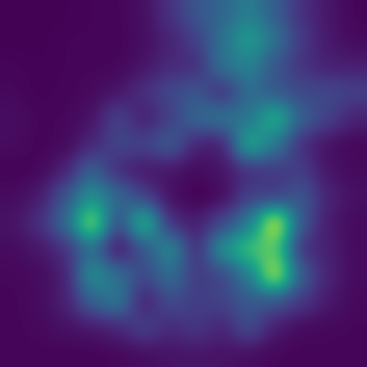}
         \end{minipage}
        }
        \subfigure[EA-AA-17]{
        \begin{minipage}[b]{0.12\linewidth}
        \includegraphics[width=\linewidth]{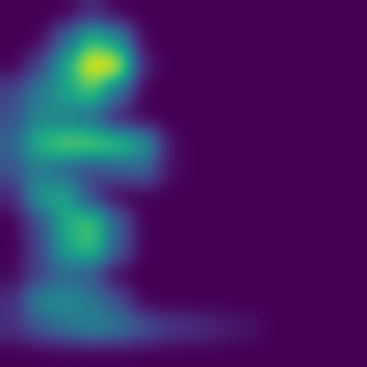}
        \includegraphics[width=\linewidth]{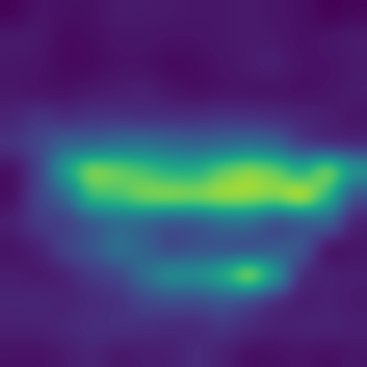}
        \includegraphics[width=\linewidth]{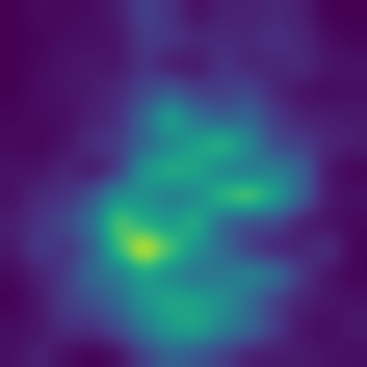}
         \end{minipage}
        }
         \subfigure[EA-AA-18]{
        \begin{minipage}[b]{0.12\linewidth}
        \includegraphics[width=\linewidth]{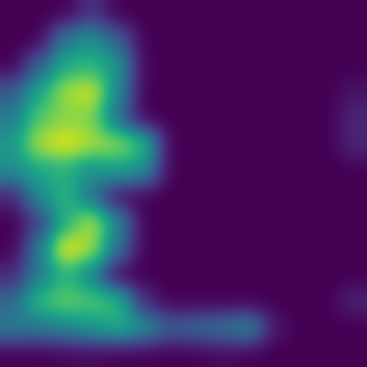}
        \includegraphics[width=\linewidth]{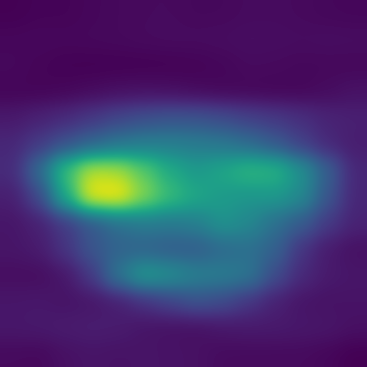}
         \includegraphics[width=\linewidth]{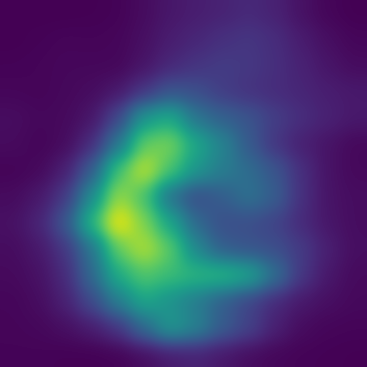}
         \end{minipage}
        }
        \caption{Visualization of exemplar attention maps from the 16th, 17th and 18th layers of AA-ResNet-34 and EA-AA-ResNet-34 models}
        \label{fig:attention_map}
 \end{figure*}


\subsection{Quality of Attention Maps}
\label{sec:analysis_attention_maps}

\begin{table}[t]
\centering
 \caption{Accuracy of attention map classification}
\scalebox{1.0}{
    \renewcommand\arraystretch{1.1}
    \begin{tabular}{lcccc}
    \toprule
     \textbf{Models} & \textbf{16th Layer} & \textbf{17th Layer} & \textbf{18th Layer} \\ 
     \midrule
     AA-ResNet-34 & 24.50 & 31.18 & 17.55 \\
     EA-AA-ResNet-34 & \textbf{31.02} & \textbf{31.71} & \textbf{31.93} \\
     \bottomrule
    \end{tabular}
}
 \label{image_analysis}
\end{table}

We analyze the attention maps in the middle layers (16th, 17th, and 18th) of the 34-layers AA-ResNet and EA-AA-ResNet architectures, and take the attention maps centered on the central pixel, because they are at an appropriate abstraction level and can better reflect the effectiveness of attention learned in the networks. The shape of each attention map is $N \times N \times K$, where $N=14$ is the image length (after pooling), and $K=8$ is the number of heads. These attention maps are taken as direct inputs to another 12-layer DenseNet~\cite{huang2017densely} model for classification. The models are trained by 30 epochs with cosine learning rate decay started by 0.05 and ended by 0.0001. If the key structures are recognized in the attention maps, we would expect a high accuracy for the classification results. Thus, we leverage the accuracy score as an indicator for the effectiveness of learned attention maps. As reported in Table \ref{image_analysis}, AA-ResNet-34 fails to induce precise attention maps in both the 16th and 18th layers. In contrast, EA-AA-ResNet-34 produces good attention maps in all three layers, showing the superiority of the evolving attention mechanism.

Figure \ref{fig:attention_map} visualizes the attention maps from a representative head of the 16th, 17th, and 18th layers for three exemplar cases of ImageNet classification. 
As illustrated, AA-ResNet prefers to extract broad and vague attention patterns. In contrast, EA-AA-ResNet generates much sharper attention maps. Meanwhile, there exists a clear evolutionary trend in three consecutive layers. For the skier case, the attention map has successfully captured the main object in the 16th layer. Then, the outline becomes much clearer at the 17th layer with the assistance of evolving attention. Finally, the 18th layer is further improved to recognize the complete skateboard. Other cases in Figure \ref{fig:attention_map} also demonstrate the privileges of evolving attention. For more details, the readers can refer to the attention maps of all heads visualized in the Appendix.

\begin{figure}[t]
	\centering
        \includegraphics[width=\linewidth]{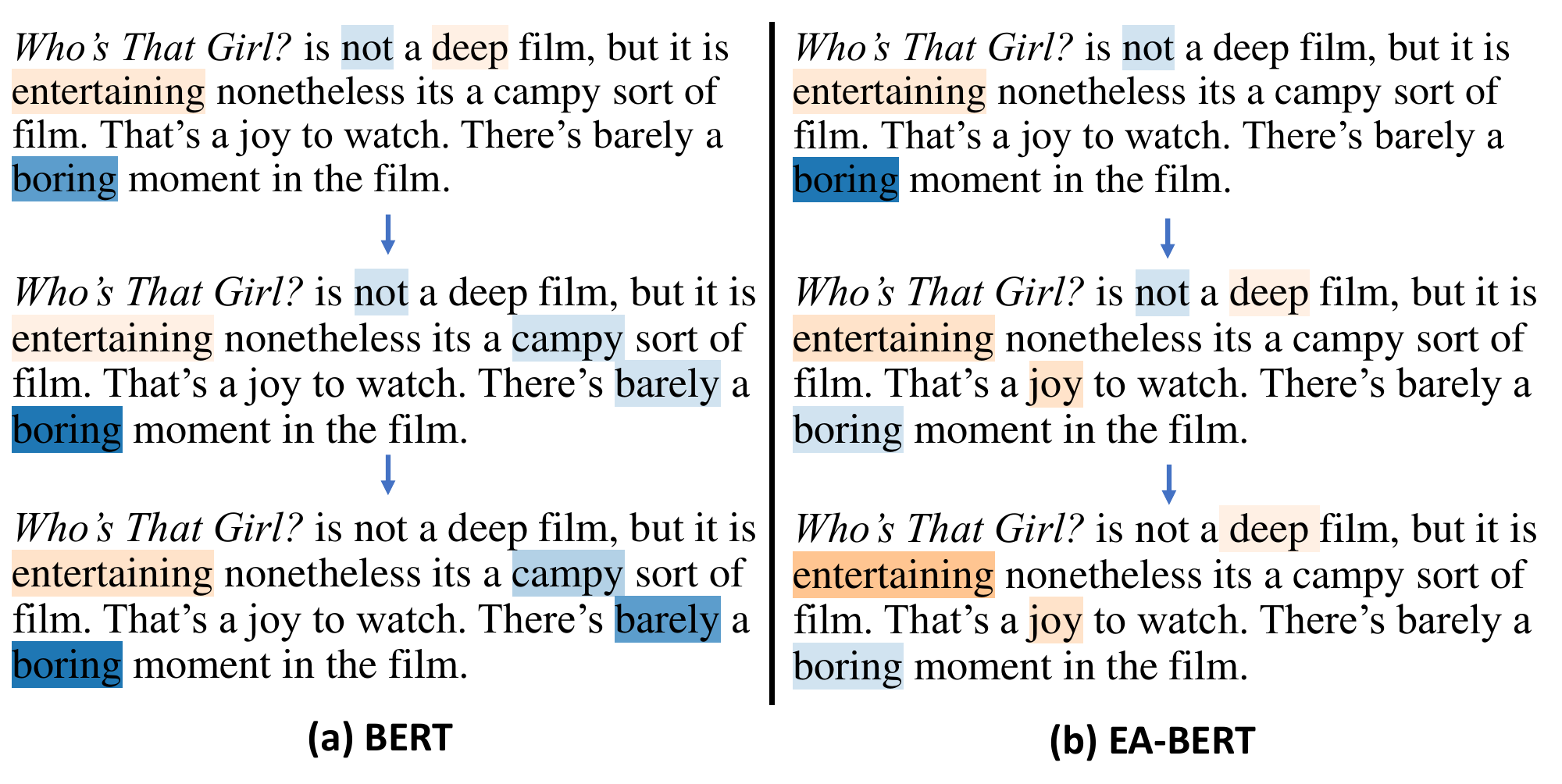}
         \caption{Comparing rationales generated by LIME for BERT and EA-BERT}
         \label{fig:lime_example}
\end{figure}


\subsection{Interpretability}
\textbf{ERASER benchmark.}
A good text representation model should not only generate correct predictions, but also give faithful reasons to explain its decisions. 
We leverage the ERASER benchmark~\cite{deyoung-etal-2020-eraser} to evaluate the interpretability of text representation models. 
The results are listed in Table \ref{app:rationale}, where `Perf." denotes accuracy (for the CoS-E dataset) or F1 score (for other datasets), AUPRC means the Area Under Precision Recall Curve; ``Comp." and ``Suff." are metrics for comprehensiveness and sufficiency, respectively, higher comprehensiveness scores and lower sufficiency scores indicate better interpretability. 
First, we utilize a text representation model that passes tokens through a BERT + LSTM encoder. Based on the text representation, we utilize both attention scores generated by vanilla \textit{Attention} and \textit{EA-Attention} for rationale generation. 
As shown in Table \ref{app:rationale}, the rationales generated by evolving attention are more accurate than those generated by the original attention mechanism.
Next, we compare the text representations obtained by BERT and EA-BERT by a state-of-the-art rationale generation method, Lime~\citep{ribeiro2016should}. Experimental results show that EA-BERT improves the performances of downstream tasks and results in better rationales.

\begin{table}[t]
\centering
 \caption{Comparison of different text representation models and rationale generation methods on ERASER benchmark. ``Perf." is accuracy (for CoS-E dataset) or F1 (for other datasets), AUPRC is Area Under the Precision Recall Curve; ``Comp." and ``Suff." stand for comprehensiveness and sufficiency, respectively.}
\scalebox{0.9}{
    \renewcommand\arraystretch{1.1}
    \begin{tabular}{lcccc}
    \toprule
     & \textbf{Perf.$\uparrow$} & \textbf{AUPRC$\uparrow$} & \textbf{Comp.$\uparrow$} & \textbf{Suff.$\downarrow$} \\ 
     \midrule
     \textbf{Movie Reviews}\\
     BERT+LSTM - Attention & 0.970 & 0.417 & 0.129 & 0.097\\
     BERT+LSTM - EA-Attention & 0.970 & \textbf{0.435} & 0.142 & \textbf{0.084}\\
     BERT+LSTM - Lime & 0.970 & 0.280 & 0.187 & 0.093\\
     EA-BERT+LSTM - Lime & \textbf{0.975} & 0.313 & \textbf{0.194} & 0.089\\
     \midrule
     \textbf{FEVER}\\
     BERT+LSTM - Attention & 0.870 & 0.235 & 0.037 & 0.122\\
     BERT+LSTM - EA-attention & 0.870 & 0.238 & 0.078 & 0.097\\
     BERT+LSTM - Lime & 0.870 & 0.291 & 0.212 & \textbf{0.014}\\
     EA-BERT+LSTM - Lime & \textbf{0.886} & \textbf{0.307} & \textbf{0.236} & \textbf{0.014}\\
     \midrule
    \textbf{MultiRC}\\
     BERT+LSTM - Attention & 0.655 & 0.244 & 0.036 & 0.052\\
     BERT+LSTM - EA-Attention & 0.655 & \textbf{0.251} & 0.054 & 0.041\\
     BERT+LSTM - Lime & 0.655 & 0.208 & 0.213 & -0.079\\
     EA-BERT+LSTM - Lime & \textbf{0.674} & 0.221 & \textbf{0.241} & \textbf{-0.089}\\
     \midrule
     \textbf{CoS-E}\\
     BERT+LSTM - Attention & 0.487 & 0.606 & 0.080 & 0.217\\
     BERT+LSTM - EA-Attention & 0.487 & \textbf{0.610} & 0.113 & 0.189 \\
     BERT+LSTM - Lime & 0.487 & 0.544 & 0.223 & 0.143\\
     EA-BERT+LSTM - Lime & \textbf{0.491} & 0.552 & \textbf{0.231} & \textbf{0.140}\\
     \midrule
     \textbf{e-SNLI}\\
     BERT+LSTM - Attention & 0.960 & 0.395 & 0.105 & 0.583\\
     BERT+LSTM - EA-Attention & 0.960 & 0.399 & 0.177 & 0.396\\
     BERT+LSTM - Lime & 0.960 & 0.513 & 0.437 & 0.389\\
     EA-BERT+LSTM - Lime & \textbf{0.969} & \textbf{0.534} & \textbf{0.445} & \textbf{0.368}\\
    \bottomrule
    \end{tabular}
}
 \label{app:rationale}
\end{table}


\noindent \textbf{Visualization of Rationales.} In Figure \ref{fig:lime_example}, we compare the rationales generated by two different models, BERT and EA-BERT, by a concrete example. The example is from the IMDB sentiment classification dataset and is misclassified to ``negative" by vanilla BERT. In the figure, the words with the blue/orange background represent rationales selected by LIME, which support the classification to ``negative"/``positive", while the darker color means the more importance a word plays. For each model, we use the outputs of the first, 6th, and the last (12th) layers as the input of the classifier to explore the changes in rationales across different layers. We found that both BERT and EA-BERT tend to classify this sample as negative when using the representation of the first layer, because negative words such as ``not" and ``boring" are highlighted. In BERT, these erroneous rationales are kept until the output of the last layer, leading to a final misclassification. In fact, the semantics expressed by these negative words can be dispelled by relative words (e.g. ``but" and ``otherwise") and adverbs (e.g. ``barely"). As shown in the right part of Figure \ref{fig:lime_example}, EA-BERT understands the inter-dependencies of these words correctly in the context, correcting the biased rationales found in the first layer, and finally find more rationales that support the classification to ``positive".


\subsection{Learning Curve Comparison}

\begin{figure}[t]\centering                                                         
\subfigure[IWSLT’14 De-En dataset]{                    
\begin{minipage}[t]{0.23\textwidth}
\centering                                                          
\includegraphics[width=\textwidth]{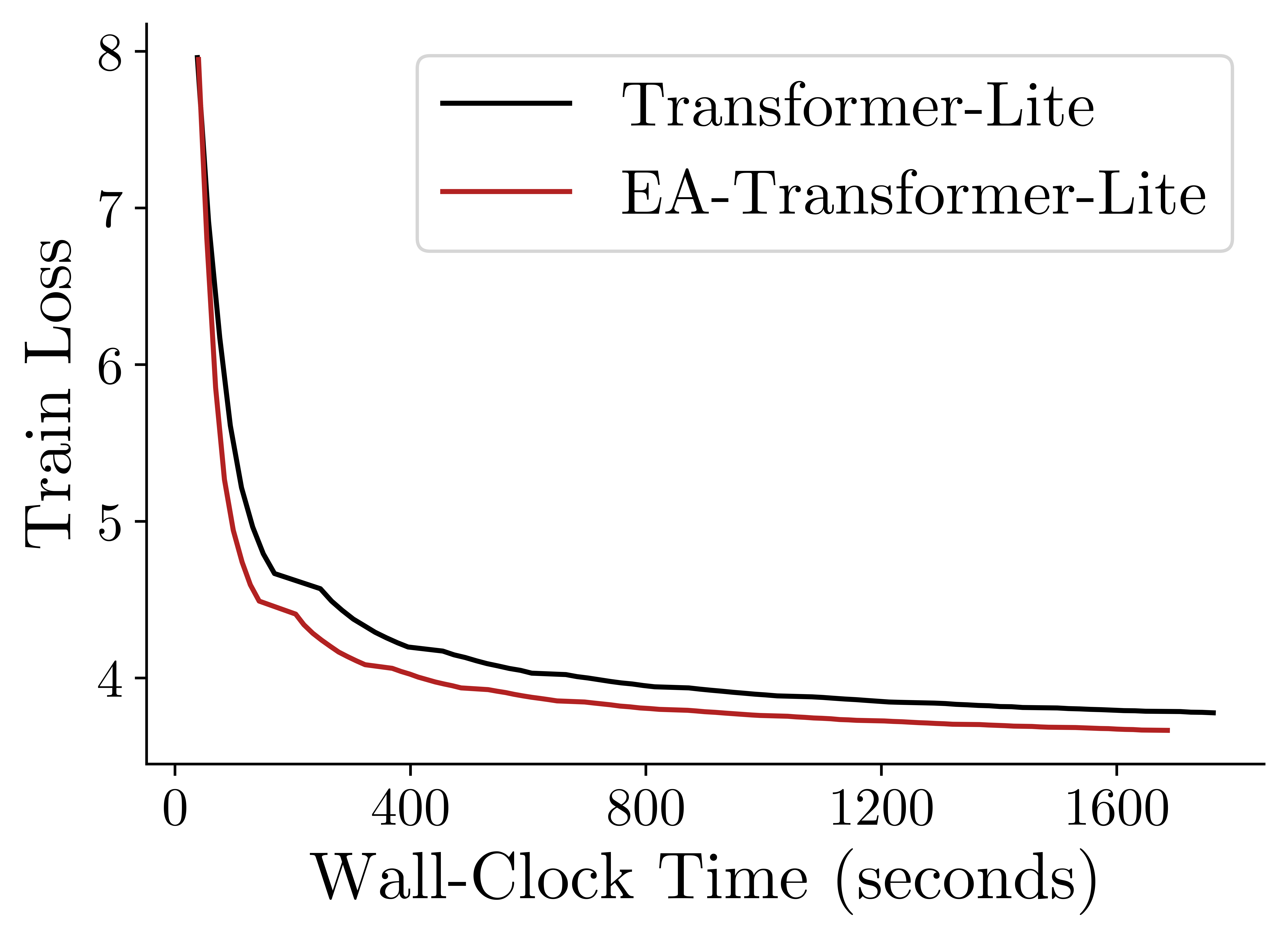}
\label{fig_lc:side:a}
\end{minipage}}\
\subfigure[AppliancesEnergy dataset]{                    
\begin{minipage}[t]{0.23\textwidth}
\centering                                                          
\includegraphics[width=\textwidth]{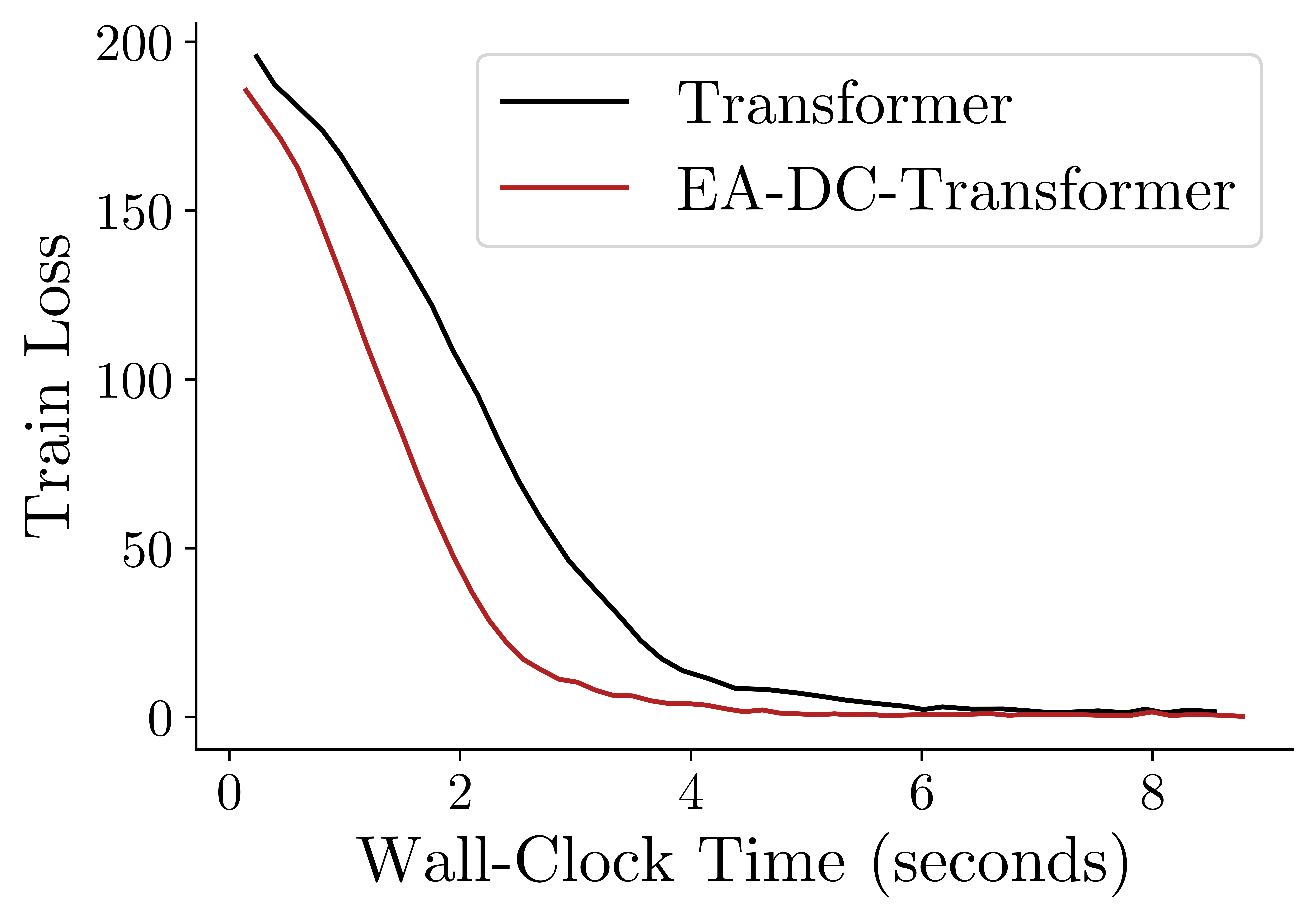}
\label{fig_lc:side:b}
\end{minipage}}
\caption{Learning curve comparison on (a) De-En machine translation dataset and (b) AppliancesEnergy time series regression dataset.} 
\label{fig:learning_curve}                                                        
\end{figure}

To demonstrate the efficiency of EA-Transformer, we further compare the learning curves of Transformer and EA-enhanced Transformer on machine translation and time series regression tasks.
As shown in Figure ~\ref{fig_lc:side:a}, EA-Transformer-Lite always achieves a lower training loss after training the same wall-clock time on the IWSLT'14 De-En dataset, although it contains relatively more 3\% FLOPs in each iteration.
At the end of the curve, EA-Transformer-Lite achieves better training loss and BLUE scores upon convergence. It can also be seen in the figure ~\ref{fig_lc:side:b} that the two-layer EA-DC-Transformer is superior to the three-layer Transformer not only in the absolute loss but also in the convergence speed. Note that EA-DC-Transformer requires fewer layers to obtain competitive results because it has an efficient convolutional layer in the evolving attention module.

\begin{figure}[t]\setcounter{subfigure}{0}
	\centering

	\subfigure[BERT \#11]{
        \begin{minipage}[t]{0.45\linewidth}
        \centering
        \includegraphics[width=\linewidth]{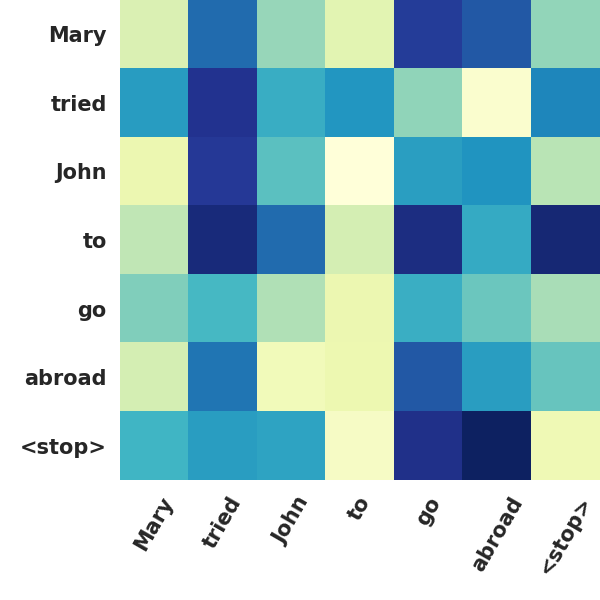}
        \label{fig:vis_raw11}
    \end{minipage}%
    }\subfigure[EA-BERT \#11]{
        \begin{minipage}[t]{0.45\linewidth}
        \centering
        \includegraphics[width=\linewidth]{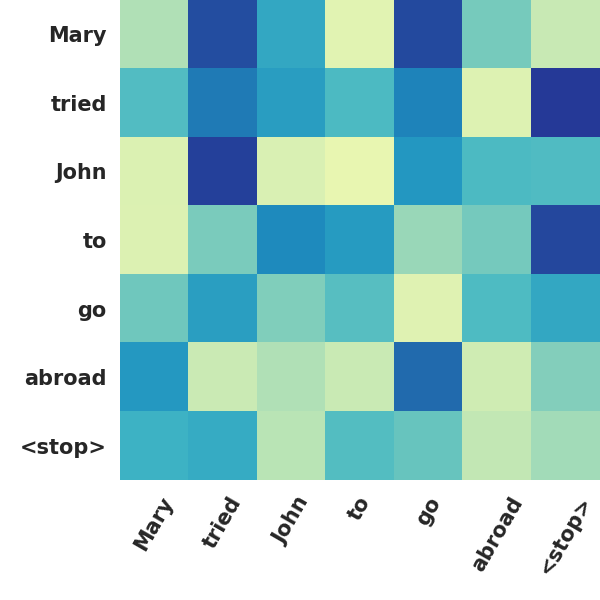}
        \label{fig:vis_conv11}
    \end{minipage}%
    }
    \subfigure[BERT \#12]{
        \begin{minipage}[t]{0.45\linewidth}
        \centering
        \includegraphics[width=\linewidth]{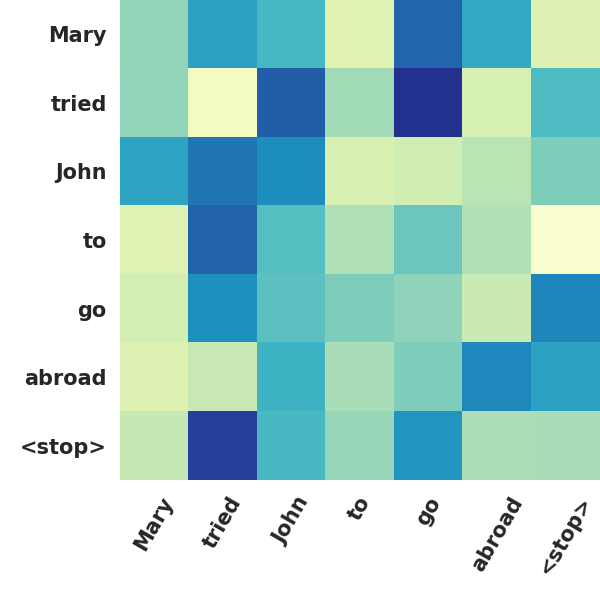}
        \label{fig:vis_raw12}
    \end{minipage}%
    }\subfigure[EA-BERT \#12]{
        \begin{minipage}[t]{0.45\linewidth}
        \centering
        \includegraphics[width=\linewidth]{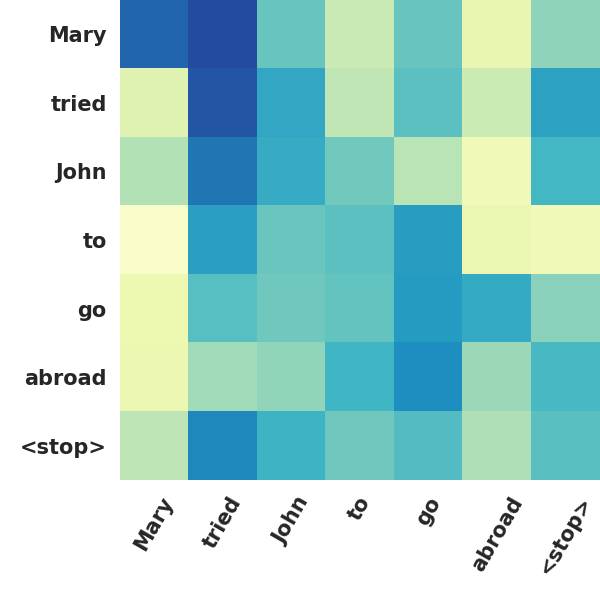}
        \label{fig:vis_conv12}
    \end{minipage}%
    }
    
    \caption{Attention maps of layer \#11 and \#12 for ``\textit{Mary tried John to go abroad.}" in BERT and EA-BERT.}
    \label{case_study_layer12}
    \label{fig:case_study}
\end{figure}

\subsection{Case Study}
\label{sec:case_study}
In order to get insight into the evolving attention mechanism, we visualize exemplar attention maps for both text and image examples and find some interesting clues.

\noindent
\textbf{Image attention.}
We compare AA-ResNet-34 and EA-AA-ResNet-34 on ImageNet classification task to understand the advantages of evolving attention for image representation. In Figure \ref{fig:attention_map}(a-g), we visualize the attention maps of layer \#16, \#17, and \#18 for AA-ResNet-34 and EA-AA-ResNet-34, respectively. From figure \ref{fig:attention_map}(a-d), we find no obvious continuity between two consecutive attention maps in AA-ResNet-34. In figure \ref{fig:attention_map}(c), the AA-ResNet-34 has obtained good detection about the object in layer-17, but the attention map becomes worse in layer-18. In contrast, EA-AA-ResNet-34, equipped with the evolving attention mechanism, keeps probing better attention maps. That is because EA-AA-ResNet-34 provides a shortcut connection and a convolution-based evolving module between adjacent layers, thus it improves the quality of attention maps as the layers go deeper.

\noindent
\textbf{Text Attention.}
We compare BERT-Base and EA-BERT-Base models on the CoLA dataset, a task of judging the grammatical correctness of a sentence. The sentence ``\textit{Mary tried John to go abroad.}" is selected for a case study. Obviously, this sentence is grammatically wrong. To give a correct prediction, the model should emphasize the phrase ``\textit{Mary tried John}" in which the grammatical error occurs.

In Figure \ref{fig:case_study}, we visualize the last attention layer (\#12) since it is closest to the output (see Figure \ref{case_study_layer12}(c-d)). For this case, the attention scores in the upper-left corner (corresponding to words ``Mary", ``tried", and 'John') determine the model prediction. We can observe that BERT-Base fails to pay enough attention to the error location, and leads to a wrong prediction. In EA-BERT-Base, the attention scores between these three words are much higher than those in BERT-Base. As a result, EA-BERT-Base correctly detects grammatical errors.
We also visualize the attention maps of the \#11 layer, which serves as input to the \#12 layer in the evolving attention mechanism. Surprisingly, we find that the attention scores of BERT-Base in the \#11 layer are more explainable than those of the \#12 layer. As shown in the upper-left corner of the attention map from BERT-Base layer \#11, it has been aware of the error location to some extent. However, due to bad evolution, the BERT-Base model fails to keep the useful knowledge in the \#12 layer and finally makes a wrong prediction. Thanks to the proposed convolution-based evolving attention mechanism, EA-BERT-Base further enhances the attention map in the \#12 layer and gives the correct prediction.

\begin{figure}[t]\setcounter{subfigure}{0}
	\centering
	\subfigure[EA-AA-ResNet-34]{
        \centering
        \includegraphics[width=0.65\linewidth]{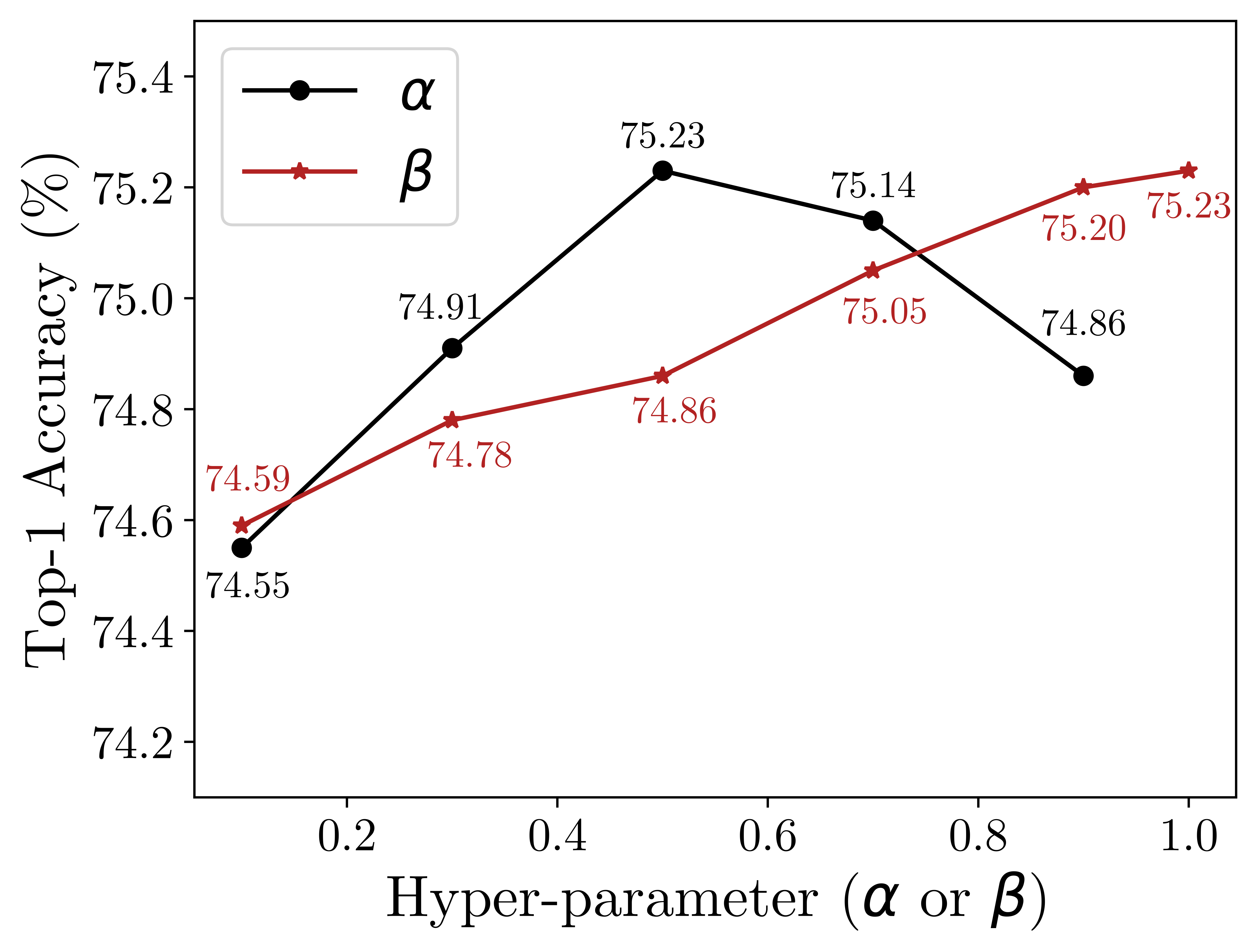}
        \label{fig:alpha}
    }
    \subfigure[EA-DC-Transformer]{
        \centering
        \includegraphics[width=0.65\linewidth]{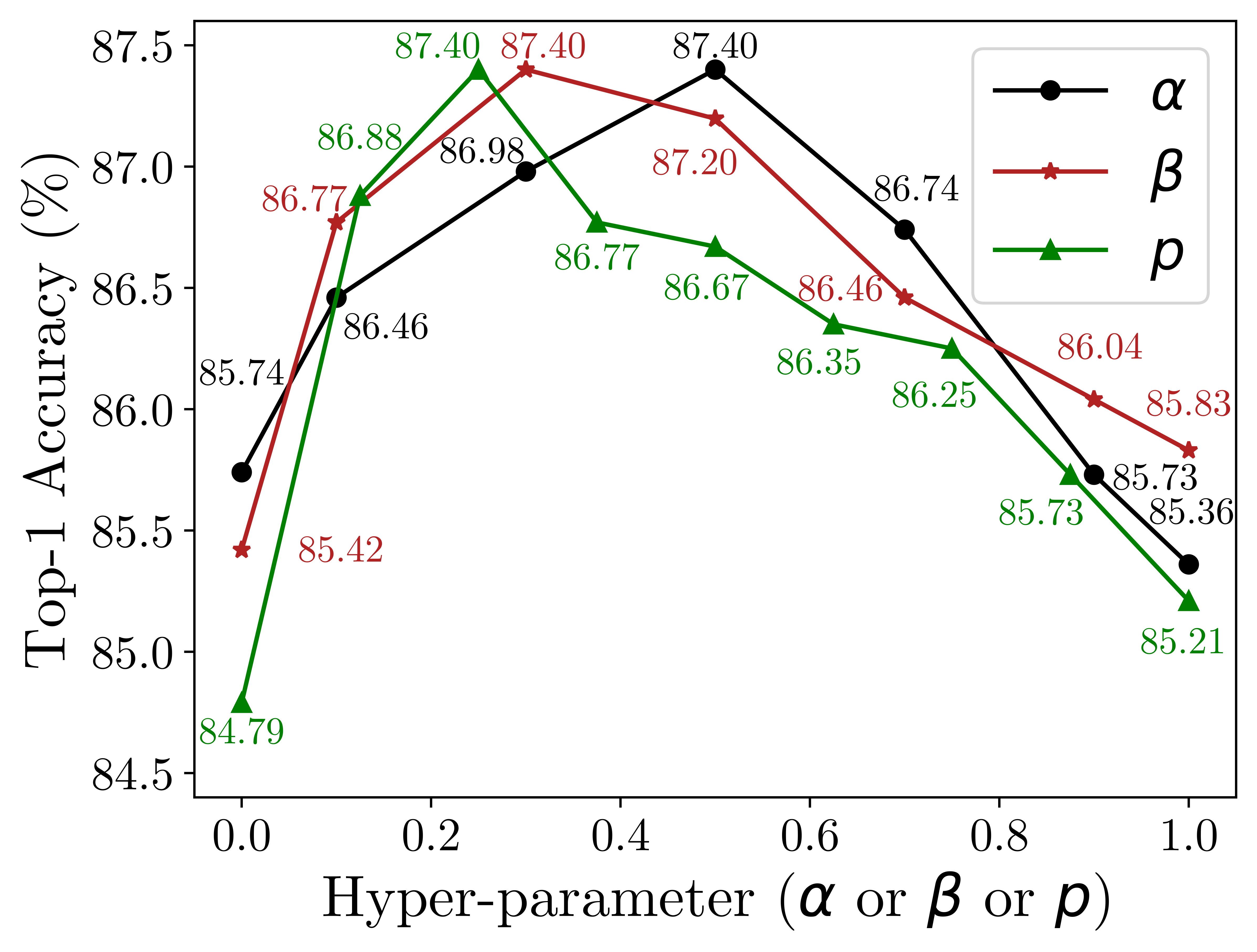}
        \label{fig:beta}
    }
    \caption{Hyper-parameter analyses}
    \label{fig:hyper-parameter}
\end{figure}


\subsection{Hyper-parameter Analysis}
\label{sec:hyper-parameter}
A hyper-parameter analysis of EA-AA-ResNet-34 is shown in Figure \ref{fig:hyper-parameter}(a). One can see that the best result is achieved at $\alpha=0.5$ and $\beta=1.0$, which significantly outperforms the vanilla transformer (equivalent to $\alpha=0$ and $\beta=0$).
We also conduct hyper-parameter analysis on the UWaveGestureLibrary dataset for time-series representation, analyzing not only $\alpha$ and $\beta$ but also $p$. 
Results are plotted in Figure~\ref{fig:hyper-parameter}(b), which shows that the best classification accuracy is achieved when $\alpha=0.5$, $\beta=0.3$, and $p=0.25$. It is worth mentioning that when $p=1$, EA-DC-Transformer will degenerate to vanilla Transformer, and when $p=0$, it will degenerate to original Dilated Convolutions. This demonstrates that EA-DC-Transformer can get a better time-series representation by organically combining dilated convolution and transformer structure.
From these two figures, we can conclude that the values of all hyper-parameters have smooth trends and are not sensitive to small fluctuations. Thus, it is feasible to perform hyper-parameter optimization on the development set and results in stable improvement on the test set.

\subsection{Limitations}
\label{sec:limitation}
The proposed convolution-based module will break the property of Transformers that the learned sequence representation is invariant to permutations of tokens. Thus, for applications that need this property, such as graph attention networks~\cite{velivckovic2018graph,wang2019heterogeneous}, we can only use $1 \times 1$ convolutions. 
Also, it is non-trivial to integrate the evolving attention mechanism to neural networks with different kinds of sparse or structured attentions~\cite{niculae2017regularized,tay2020sparse,dao2022monarch}, while how to combine their advantages remains for future study. 
\section{Conclusion}
In this paper, we propose a novel mechanism, Evolving Attention, which is applicable to various kinds of attention neural networks. Equipped with this mechanism, the Convolution-enhanced Evolving Attention Networks produce better attention maps and achieve superior performance on various tasks in time-series, NLP, and CV domains.
Future works are considered in three aspects. First, we will apply the evolving attention networks to more tasks and domains, such as image generation~\cite{parmar2018image}, multi-modal tasks~\cite{khan2021transformers}, recommendation systems~\cite{sun2019bert4rec}, and graph neural networks~\cite{defferrard2016convolutional}. Second, we would like to investigate other modules instead of convolutions to capture the generic patterns in attention maps. Third, we plan to explore bi-directional decoders that further leverage the power of convolutions.


%



\ifCLASSOPTIONcompsoc
 \section*{Acknowledgments}
\else
  \section*{Acknowledgment}
\fi
The authors would like to thank Dr. Zhouchen Lin and anonymous reviewers for their valuable suggestions to polish this paper.
This work was supported by the National Key Research and Development Program of China (No. 2020YFB2103402). GH is supported in part by
Beijing Academy of Artificial Intelligence.

\ifCLASSOPTIONcaptionsoff
  \newpage
\fi



%
{
\small
\bibliographystyle{IEEEtran}
\bibliography{reference}
}



%





\clearpage

\end{document}